\documentclass[journal]{IEEEtran}

\usepackage{times}
\usepackage{epsfig}
\usepackage{graphicx}
\usepackage{amsmath}
\usepackage{amssymb}
\usepackage{makecell}
\usepackage{todonotes}
\usepackage{tablefootnote}
\usepackage{longtable}
\usepackage{caption}
\usepackage{enumitem}
\usepackage[symbol]{footmisc}
\usepackage{comment}
\usepackage{adjustbox}
\usepackage[switch,columnwise]{lineno}
\usepackage[normalem]{ulem}
\RequirePackage{color}\definecolor{RED}{rgb}{1,0,0}\definecolor{BLUE}{rgb}{0,0,1} 

\newcommand{\RNum}[1]{\uppercase\expandafter{\romannumeral #1\relax}}
\newcommand{\etal}{\emph{et al.}}

\providecommand{\DIFaddbegin}{} 
\providecommand{\DIFaddend}{} 
\providecommand{\DIFdelbegin}{} 
\providecommand{\DIFdelend}{} 
\providecommand{\DIFaddbeginFL}{} 
\providecommand{\DIFaddendFL}{} 
\providecommand{\DIFdelbeginFL}{} 
\providecommand{\DIFdelendFL}{} 

\RequirePackage[normalem]{ulem} 
\RequirePackage{color}\definecolor{RED}{rgb}{1,0,0}\definecolor{BLUE}{rgb}{0,0,1} 
\providecommand{\DIFaddbegin}{} 
\providecommand{\DIFaddend}{} 
\providecommand{\DIFdelbegin}{} 
\providecommand{\DIFdelend}{} 
\providecommand{\DIFaddbeginFL}{} 
\providecommand{\DIFaddendFL}{} 
\providecommand{\DIFdelbeginFL}{} 
\providecommand{\DIFdelendFL}{} 
\newcommand{\DIFscaledelfig}{0.5}
\RequirePackage{settobox} 
\RequirePackage{letltxmacro} 
\newsavebox{\DIFdelgraphicsbox} 
\newlength{\DIFdelgraphicswidth} 
\newlength{\DIFdelgraphicsheight} 
\LetLtxMacro{\DIFOincludegraphics}{\includegraphics} 
\newcommand{\DIFaddincludegraphics}[2][]{{\color{blue}\fbox{\DIFOincludegraphics[#1]{#2}}}} 
\newcommand{\DIFdelincludegraphics}[2][]{
	\sbox{\DIFdelgraphicsbox}{\DIFOincludegraphics[#1]{#2}}
	\settoboxwidth{\DIFdelgraphicswidth}{\DIFdelgraphicsbox} 
	\settoboxtotalheight{\DIFdelgraphicsheight}{\DIFdelgraphicsbox} 
	\scalebox{\DIFscaledelfig}{
		\parbox[b]{\DIFdelgraphicswidth}{\usebox{\DIFdelgraphicsbox}\\[-\baselineskip] \rule{\DIFdelgraphicswidth}{0em}}\llap{\resizebox{\DIFdelgraphicswidth}{\DIFdelgraphicsheight}{
				\setlength{\unitlength}{\DIFdelgraphicswidth}
				\begin{picture}(1,1)
				\thicklines\linethickness{2pt} 
				{\color[rgb]{1,0,0}\put(0,0){\framebox(1,1){}}}
				{\color[rgb]{1,0,0}\put(0,0){\line( 1,1){1}}}
				{\color[rgb]{1,0,0}\put(0,1){\line(1,-1){1}}}
				\end{picture}
			}\hspace*{3pt}}} 
} 
\LetLtxMacro{\DIFOaddbegin}{\DIFaddbegin} 
\LetLtxMacro{\DIFOaddend}{\DIFaddend} 
\LetLtxMacro{\DIFOdelbegin}{\DIFdelbegin} 
\LetLtxMacro{\DIFOdelend}{\DIFdelend} 
\DeclareRobustCommand{\DIFaddbegin}{\DIFOaddbegin \let\includegraphics\DIFaddincludegraphics} 
\DeclareRobustCommand{\DIFaddend}{\DIFOaddend \let\includegraphics\DIFOincludegraphics} 
\DeclareRobustCommand{\DIFdelbegin}{\DIFOdelbegin \let\includegraphics\DIFdelincludegraphics} 
\DeclareRobustCommand{\DIFdelend}{\DIFOaddend \let\includegraphics\DIFOincludegraphics} 
\LetLtxMacro{\DIFOaddbeginFL}{\DIFaddbeginFL} 
\LetLtxMacro{\DIFOaddendFL}{\DIFaddendFL} 
\LetLtxMacro{\DIFOdelbeginFL}{\DIFdelbeginFL} 
\LetLtxMacro{\DIFOdelendFL}{\DIFdelendFL} 
\DeclareRobustCommand{\DIFaddbeginFL}{\DIFOaddbeginFL \let\includegraphics\DIFaddincludegraphics} 
\DeclareRobustCommand{\DIFaddendFL}{\DIFOaddendFL \let\includegraphics\DIFOincludegraphics} 
\DeclareRobustCommand{\DIFdelbeginFL}{\DIFOdelbeginFL \let\includegraphics\DIFdelincludegraphics} 
\DeclareRobustCommand{\DIFdelendFL}{\DIFOaddendFL \let\includegraphics\DIFOincludegraphics} 

\begin{document}
	%
	\title{Multi-Scale Thermal to Visible Face Verification via Attribute Guided Synthesis}

	\author{Xing Di,~\IEEEmembership{Student Member,~IEEE} and Benjamin S. Riggan,~\IEEEmembership{Member,~IEEE} and Shuowen Hu,~\IEEEmembership{Member,~IEEE} and Nathaniel J. Short,~\IEEEmembership{Member,~IEEE}
		Vishal~M.~Patel,~\IEEEmembership{Senior Member,~IEEE}
		\thanks{Xing Di is with the Whiting School of Engineering, Johns Hopkins University, 3400 North Charles Street, Baltimore, MD 21218-2608, e-mail: xing.di@jhu.edu}
		\thanks{Benjamin S. Riggan is with the University of Nebraska, e-mail: briggan2@unl.edu}
		\thanks{Shuowen Hu is with the U.S. Army DEVCOM Army Research Laboratory (ARL), e-mail: shuowen.hu.civ@mail.mil}
		\thanks{Nathaniel J. Short is with Booz Allen Hamilton, e-mail:short\_nathaniel@bah.com}
		\thanks{Vishal M. Patel is with the Whiting School of Engineering, Johns Hopkins University, e-mail: vpatel36@jhu.edu}
	}

	\markboth{Journal of \LaTeX\ Class Files,~Vol.~xx, No.~x, Month~2017}%
	{Shell \MakeLowercase{\textit{et al.}}: Bare Demo of IEEEtran.cls for IEEE Journals}
	%
	
	\maketitle
	

	\begin{abstract}
		Thermal-to-visible face verification is a challenging problem due to the large domain discrepancy between the modalities.  Existing approaches either attempt to synthesize visible faces from thermal faces or learn domain-invariant robust features from these modalities for cross-modal matching. In this paper, we use attributes extracted from visible images to synthesize attribute-preserved visible images from thermal imagery for cross-modal matching. A pre-trained attribute predictor network is used to extract the attributes from the visible image. Then, a novel multi-scale generator is proposed to synthesize the visible image from the thermal image guided by the extracted attributes. Finally, a pre-trained VGG-Face network is leveraged to extract features from the synthesized image and the input visible image for verification.  Extensive experiments evaluated on three datasets (ARL Face Database, Visible and Thermal Paired Face Database, and Tufts Face Database) demonstrate that the proposed method achieves state-of-the-art  performance. In particular, it achieve around 2.41\%, 2.85\% and 1.77\% improvements in Equal Error Rate (EER) over the state-of-the-art methods on the ARL Face Database, Visible and Thermal Paired Face Database, and Tufts Face Database, respectively. An extended dataset (ARL Face Dataset volume III) consisting of polarimetric thermal faces of 121 subjects is also introduced in this paper.	Furthermore, an ablation study is conducted to demonstrate the effectiveness of different modules in the proposed method.
		
	\end{abstract}


	\
	
	\section{Introduction} \label{introduction}
	
	Face Recognition (FR) is one of the most widely studied problems in computer vision and biometrics research communities due to its applications in authentication, surveillance, and security.  Various methods have been developed over the last two decades that specifically attempt to address the challenges such as aging, occlusion, disguise, variations in pose, expression, and illumination.  In particular, convolutional neural network (CNN) based FR methods have gained significant traction in recent years \cite{FR_SPM_2018}.  This is mainly due to the availability of large annotated datasets, affordability of graphics processing units (GPUs), and trainability of nonlinear layers of deep neural networks employing activations functions (e.g., ReLU, ELU ) that alleviated issues with diminishing/exploding gradients.  Many deep CNN-based methods \cite{parkhi2015deep,schroff2015facenet,wen2016discriminative,JC_WACV2016,FR_SPM_2018, HyperFace,deng2019arcface, wu2018light} have achieved state-of-the-art performances on various FR benchmarks.
	
	\begin{figure}[t!]
		\centering
		\includegraphics[width=1.00\linewidth]{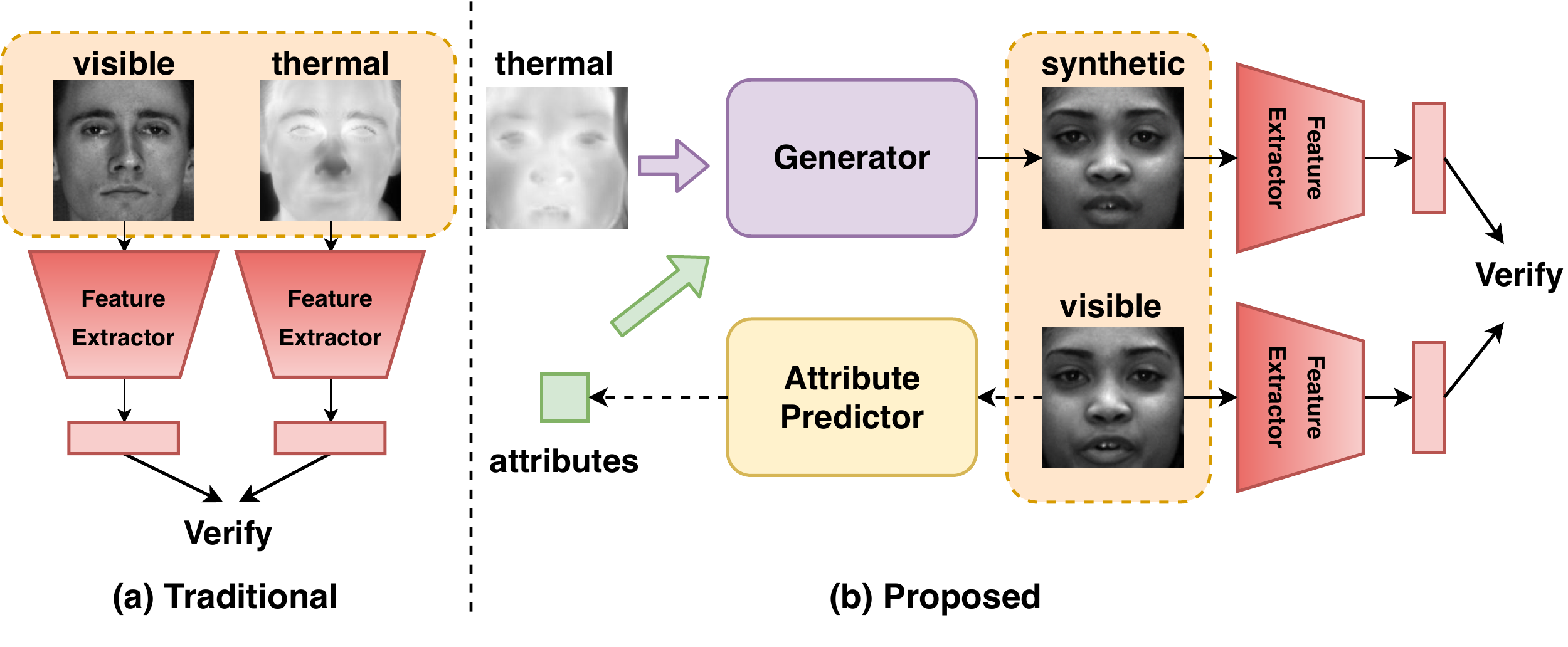}
		\caption{(a) Traditional heterogeneous face verification approaches use the features directly extracted from different modalities for verification \cite{hu2015thermal,klare2010heterogeneous,thermalfacerecognition2012,studymidwave2012}. (b) The proposed heterogeneous face verification approach uses a thermal face and semantic attributes to synthesize a visible face. Then, deep features extracted from the synthesized and visible faces are used for verification.}
		\label{fig:framework}
	\end{figure}

	
	\begin{figure*}
		\centering
		\includegraphics[width=0.95\linewidth]{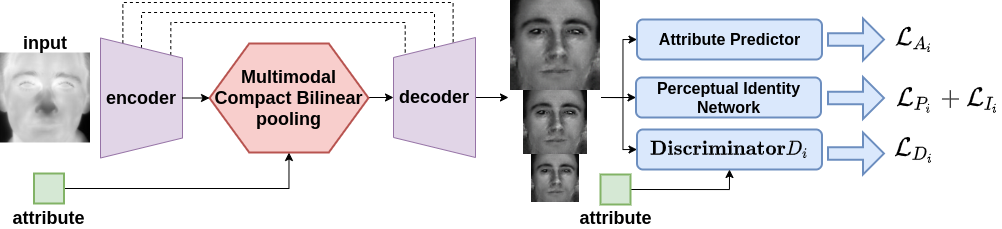}\\
		\raggedleft
		\caption{ A single generator with multi-scale resolution output is proposed to synthesize high-quality images by leveraging hierarchical information at different scales. Multimodal Bilinear Pooling (MCB) pooling is proposed to fuse the semantic attribute information with the image feature in the latent space.
			In order to make sure that the synthesized image maintains the identity and semantic attributes, a multi-purpose objective function is adopted which consists of adversarial loss $\mathcal{L}_{D_{i}}$,  $\mathcal{L}_{1}$ loss, perceptual loss $\mathcal{L}_{P_{i}}$, identity loss $\mathcal{L}_{I_{i}}$ and  attribute preserving loss $\mathcal{L}_{A_{i}}$.}
		\label{fig:framework3}
	\end{figure*}

	Despite the success of CNN-based methods in addressing various challenges in FR, they are fundamentally limited to recognizing
	face images that are collected near-infrared spectrum.  In many practical scenarios such
	as surveillance in low-light conditions, one has to detect and recognize faces that are captured using thermal modalities  \cite{hu2016polarimetric,riggan2016estimation,short2015improving,zhang2017generative,Riggan2018thermal,klare2010heterogeneous,nicolo2012long, lezama2017not, bourlai2010cross, BOURLAI201614}.  However, the performance of many deep learning-based methods degrades significantly when they are presented with thermal face images.  For example, it was shown in \cite{zhang2017generative,Riggan2018thermal,di2018polarimetric,di2019icb} that simply using deep features extracted from both thermal and visible facial images are not sufficient enough for heterogeneous face recognition.  The performance degradation is mainly due to the significant distributional change between the thermal and visible domains as well as a lack of sufficient data for training the deep networks for cross-modal synthesis and matching.
	
	Several attempts have been made to address the thermal-to-visible cross-spectrum FR problem \cite{Riggan2018thermal,riggan2016estimation,zhang2017generative,di2018polarimetric,zhang2018synthesis}.  Riggan \etal   \cite{riggan2016estimation} proposed a two-step method
	(visible feature estimation and visible image reconstruction) to solve the heterogeneous FR problem. Zhang \etal \cite{zhang2017generative} proposed a generative adversarial network (GAN) based method that fuses different Stokes images to synthesize a visible face image given the corresponding polarimetric thermal images.  Recently, Riggan \etal \cite{Riggan2018thermal} developed a global and local region-based method to improve the discriminative quality of the synthesized visible imagery.  Recently, Zhang \etal \cite{zhang2018synthesis} introduced a multi-stream feature-level fusion method to synthesize high-quality visible images from polarimetric thermal images. Though these methods are able to synthesize photo-realistic visible face images to some extent, the synthesized results in  \cite{zhang2017generative,reed2016generative,Riggan2018thermal} are still far from optimal and they tend to lose some semantic attribute information such as expression, facial hair, gender, etc. Such reconstructions may degrade the performance of thermal-to-visible face verification.

	In this paper, we take a different approach to the problem of thermal-to-visible matching.  Fig.~\ref{fig:framework} compares the traditional cross-modal verification problem with that of the proposed attribute-preserved heterogeneous face verification approach.  Given a visible and thermal image pair, the traditional approach first extracts some features from these images and then verifies the identity based on the extracted features \cite{klare2010heterogeneous} (see Fig.~\ref{fig:framework}(a)).  In contrast,  we propose a novel framework in which we make use of the attributes extracted from the visible image to synthesize the attribute-preserved visible image from the input thermal image for matching (see Fig.~\ref{fig:framework}(b)).  In particular, a pre-trained VGG-Face model \cite{parkhi2015deep} is used to extract the attributes from the visible image.  Then, a novel Multi-Scale Attribute Preserved Generative Adversarial Network (Multi-AP-GAN) is proposed to synthesize the visible image from the thermal image guided by the extracted attributes.  Finally, a pre-trained VGG-Face network is used to extract features from the synthesized and the input visible images for verification.

	The proposed Multi-AP-GAN model is inspired by the recent works \cite{lee2015deeply,zhang2017mdnet,xie2015holistically,zhao2017pyramid,zhang2018photographic}, in which deep supervision  \cite{lee2015deeply} is used at intermediate convolutional layers to learn better feature representations. Specifically, the Multi-AP-GAN consists of two parts: (i) a multimodal compact bilinear (MCB) pooling-based generator \cite{fukui2016multimodal,gao2016compact}, and (ii) a generator with the multi-scale architecture.  The MCB pooling module fuses the given attributes with the image features.  The multi-scale architecture aims to improve the synthesis image quality by leveraging hierarchical representations of CNNs at different image resolutions.

	Fig.~\ref{fig:framework3} provides an overview of the proposed  Multi-AP-GAN framework. A single generator with a series of distinct discriminators are employed to learn the multi-scale adversarial discrimination at different scales \cite{wang2018high}.  The generator fuses the extracted attribute vector with the image feature vector in the latent space.  On the other hand, each discriminator uses triplet pairs (real image/true attributes, fake image/true attributes, fake image/wrong attributes) to not only discriminate between real and fake images but also to discriminate between the image and the attributes.  In order to generate high-quality and attribute-preserved images, the generator is optimized by a multi-purpose objective function consisting of adversarial loss \cite{goodfellow2014generative},  $L_{1}$ loss, perceptual loss \cite{johnson2016perceptual},  identity loss \cite{zhang2017generative} and  attribute preserving loss.

	To summarize, this paper makes the following contributions:
	\begin{itemize}
		
		\item  We propose a novel thermal-to-visible face verification framework in which Multi-AP-GAN is developed for synthesizing high-quality visible faces from thermal images guided by facial attributes.
		
		\item We propose a single generator with a multi-scale output architecture and a Multimodal Compact Bilinear (MCB) pooling module \cite{fukui2016multimodal,gao2016compact} to generate high-quality visible images.

		\item A novel triplet-pair discriminator is proposed,  where the discriminator \cite{reed2016generative} not only learns to discriminate between real/fake images as well as images/visual-attributes.
		
		\item An extended version of the ARL polarimetric thermal face database consisting of data from 121 individuals is introduced in this work.

		\item      Extensive experiments are conducted on three different volumes of the ARL Multimodal Facial Database \cite{hu2016polarimetric,zhang2018synthesis} as well as the Thermal and Visible Paired Face Database \cite{Mallat18}, and  comparisons  are  performed  against  several  recent  state-of-the-art
		approaches. Furthermore, an ablation study is conducted to demonstrate the improvements obtained by including semantic attribute information for synthesis.  	
		
	\end{itemize}
	
	Note that the proposed Multi-AP-GAN framework can be viewed as an extended version of our earlier paper in the 2018 BTAS proceedings \cite{di2018polarimetric}.  However, the generators used in both papers are quite different.  The generator in \cite{di2018polarimetric} is a single-scale generator whereas a multi-scale generator is proposed in this paper.  Furthermore, a new polarimetric thermal face dataset consisting of multimodal data from 121 subjects is introduced in this paper.  Extensive experiments and analysis are presented using the new dataset as well as the Thermal and Visible Paired Face Database \cite{Mallat18}.

	The rest of the paper is organized as follows. In Section \ref{sec: related work}, we review a few related works on visible to thermal face synthesis and matching. Details of the proposed Multi-AP-GAN method are given in Section \ref{sec: proposed method}. Datasets and corresponding protocols are described in Section \ref{sec: dataset}.  Experimental results are presented in Section \ref{sec: experimental result}. Finally, Section \ref{sec: conclusion} concludes the paper with a brief summary and discussion.

	\section{Related Work} \label{sec: related work}
	In this section, we review some related works on thermal-to-visible face synthesis and recognition. 
	
	\subsection{Feature-based Thermal-Visible Face Recognition}
	Traditional thermal-to-visible face verification methods first extract features from the visible and thermal images and then verify the identity based on the extracted features (See Fig.~\ref{fig:framework}).  Both hand-crafted and learned features have been investigated in the literature. Buddharaju \etal~\cite{buddharaju2007physiology} proposed a method that leverages physiological information based on the superficial blood vessel network for face recognition in themral imagery.  In \cite{wesley2012comparative} Wesley \etal presented a comparative analysis of performance of automated facial expression recognition from thermal videos, visual facial videos, and their fusion using  principal component analysis (PCA) based features.  Gyaourova \etal~ \cite{gyaourova2004fusion} proposed a multimodal fusion method by combining information from both thermal and visible images for face recognition.  Hu \etal \cite{hu2015thermal} proposed a partial least squares (PLS) regression-based approach for heterogeneous face matching. Klare \etal \cite{klare2013heterogeneous} developed a generic framework for cross-modal FR based on kernel prototype nonlinear similarities.   Another multiple texture descriptor fusion-based method was proposed by Bourlai \etal in \cite{studymidwave2012} for cross-modal FR. In \cite{thermalfacerecognition2012} PLS-based discriminant analysis approaches were used to correlate the thermal face images to the visible face signatures. Gurton \etal \cite{Gurton:14} and Nathaniel \etal \cite{short2015exploiting,Short:15}  proposed to use the polarization-state information of thermal emissions to enhance the performance of thermal FR.   Wu \etal \cite{wu2019disentangled} introduced a disentangled variational representation for crossmodal matching in which a face representation is modeled with an intrinsic identity information and its within-person variations. He \etal \cite{he2017learning} proposed a network which maps both NIR and VIS images to a compact Euclidean space for matching.   Later on, they added more constraints on the representation by utilizing Wassertain distance \cite{he2018wasserstein} and adversarial learning \cite{he2019adversarial}. Fu \etal~ \cite{fu2019dual} proposed a framework which generates new paired images with abundant intra-class diversity to reduce the domain gap of heterogeneous face recognition.

	\subsection{Synthesis-based Thermal-Visible Face Recognition}
	Synthesis-based thermal-to-visible face verification algorithms leverage the synthesized visible faces for verification.  Due to the success of CNNs and recently introduced GANs in synthesizing realistic images, various deep learning-based approaches have been proposed in the literature for thermal-to-visible face synthesis \cite{Riggan2018thermal,zhang2017generative,zhang2017tv,riggan2016estimation,he2019adversarial,yu2019lamp}. For instance, Riggan \etal \cite{riggan2016estimation} proposed a two-step procedure
	(visible feature estimation and visible image reconstruction) to solve the cross-modal verification problem. Zhang \etal \cite{zhang2017generative} proposed an end-to-end GAN-based approach for synthesizing photo-realistic visible face images from the corresponding polarimetric thermal images. Recently Riggan \etal \cite{Riggan2018thermal} proposed a new synthesis method to enhance the discriminative quality of generated visible face images by leveraging both global and local facial regions.Zhang \etal \cite{zhang2018synthesis} introduced a multi-stream fusion-based generative model for cross-modal face verification. Di \etal \cite{di2018polarimetric} proposed a GAN-based network called AP-GAN to improve the synthesized visible image by utilizing visual attributes. Di \etal \cite{di2019icb} proposed another unsupervised generative model which combines features from both thermal-to-visible and visible-to-thermal synthesized images for heterogeneous face verification. Recently Pereira \etal~ \cite{de2018heterogeneous}  proposed a generic adaptation-based network for heterogeneous face recognition. He \etal \cite{he2019adversarial} 
	proposed a generative model for thermal-to-visible face synthesis by utilizing texture inpainting and pose correction. Another improved FusionNet was proposed in \cite{litvin2019novel}, which increases robustness against overfitting using dropout for a thermal-to-visible generation.  This method was evaluated on the RGB-D-T dataset \cite{nikisins2014rgb}. 	Recently, Mallat \etal \cite{mallat2019cross,damer2019cascaded} proposed a cascaded model which is optimized by the contextual loss \cite{mechrez2018contextual} for cross-spectrum synthesis.  An attribute-guided visible face synthesis method using a conditional CycleGAN framework was proposed in \cite{lu2018attribute}.

	\section{Proposed Method} \label{sec: proposed method}
	
	\begin{figure*}[t]
		\centering
		\includegraphics[width=0.9\linewidth]{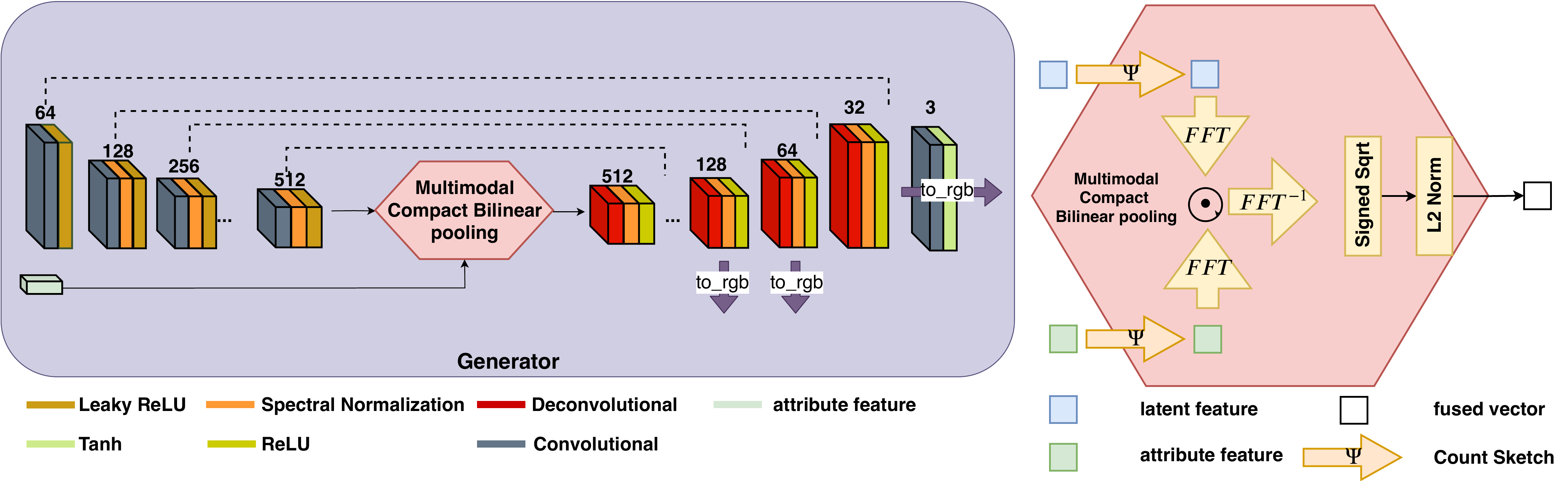} 
		\vskip 0mm \hskip 45mm (a) The multi-scale generator \hskip 15mm (b) Multimodal Compact Bilinear (MCB) pooling 
		\caption{ The network architecture of multi-scale generator and multimodal compact bilinear (MCB) pooling in details.}
		\label{fig:generator}
	\end{figure*}
	
	\begin{figure}[t]
		\centering
		\includegraphics[width=\linewidth]{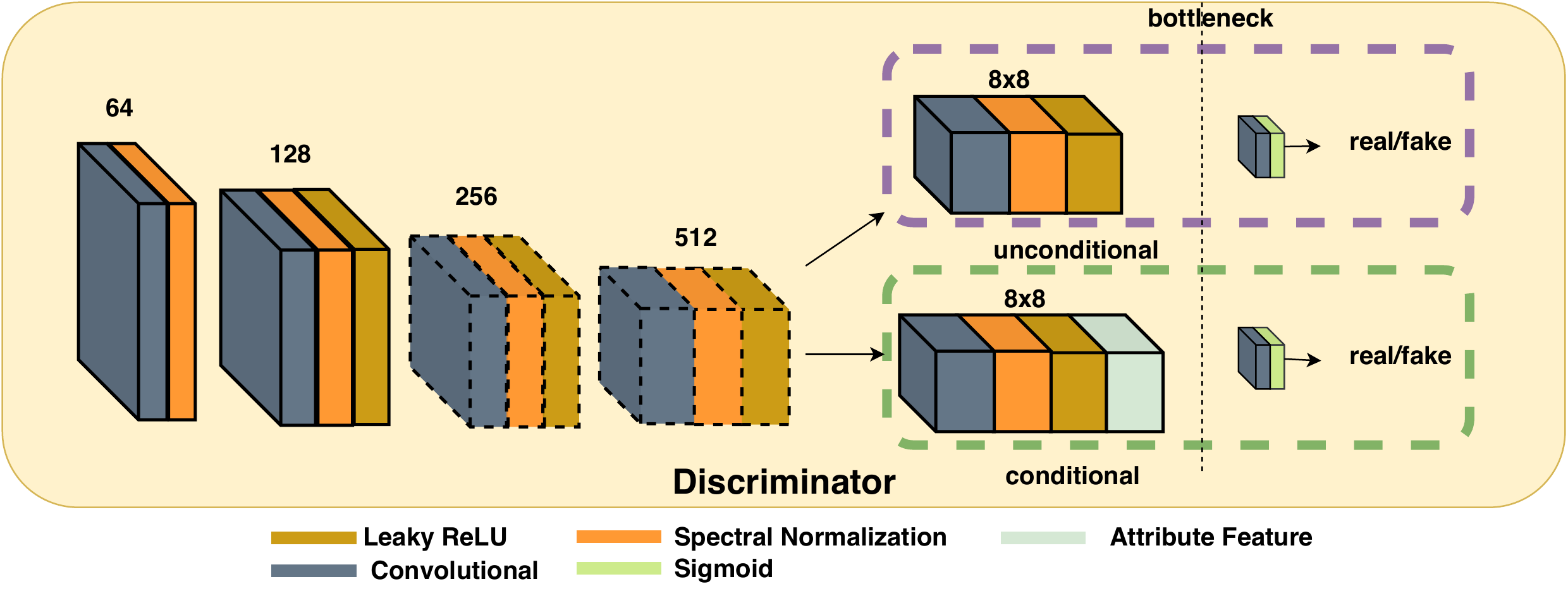}
		\caption{An overview of the triplet-pair-input discriminator. The triplet-pair-input discriminator is composed of a conditional and an unconditional streams. The unconditional stream aims to discriminate the fake and real images. The conditional stream aims to discriminate between the image and the corresponding attributes. In order to keep the bottleneck feature map size to be consistent to $8\times 8$ for different input image resolution scale, a different number of downsampling layers (dash-line cubic) are utilized. }
		\label{fig:discriminator}
	\end{figure}

	In this section, we discuss the details of the proposed Multi-AP-GAN method (see Fig.~\ref{fig:framework3}). In particular, we discuss the proposed attribute predictor, multi-scale generator, a series of distinct accompanying discriminators and the loss function used to train these networks.

	\subsection{Attribute Predictor} \label{attribute predictor}
	To efficiently extract attributes from a given visible face, an attribute predictor is fine-tuned based on the VGG-Face network \cite{parkhi2015deep} using the annotated attributes. This network is trained separately from Multi-AP-GAN. The fine-tuned network is used in both obtaining the visible face attributes and for capturing the attribute loss when training the generator and discriminator. When fine-tuning the network, a binary cross-entropy loss is used and the final fully-connected layer has the same dimension as the number of visual attributes. The predictor is selected based on the lowest loss error.

	\subsection{Generator} \label{sec: generator}
	A  U-net structure \cite{ronneberger2015u} is used as the building block for the multi-scale generator since it is able to better capture the large receptive field and also able to efficiently address the vanishing gradient problem. In addition,  to effectively combine the extra facial attribute information into the building block, we fuse the attribute vector and the image feature in the latent space \cite{reed2016generative,zhang2017generative,di2017face}. Note that the attributes are extracted from the given visible face using the fine-tuned model as discussed above. The generator architecture is illustrated in Fig~\ref{fig:generator}(a).
	
	In our experiments, we observe that simple concatenation of the two vectors (encoded image vector and attribute vector) does not work well. One possible reason is that both vectors are significantly different in terms of their dimensionality.   Thus, we adopt the well-known MCB pooling method \cite{fukui2016multimodal,gao2016compact} to overcome this issue. Instead of simple concatenation, MCB leverages the following two techniques: bilinear pooling and sketch count. Bilinear pooling is the outer-product and linearization of two vectors, where all elements of both vectors are interacting with each other in a multiplicative way. In order to overcome the high-dimension computation of bilinear pooling, Pham \etal \cite{pham2013fast} implemented the count sketch of the outer product of two vectors, which involves the Fast Fourier Transform ($FFT$) and inverse Fast Fourier Transform ($FFT^{-1}$). The architecture of the MCB module is shown in Fig~\ref{fig:generator}(b).
	
	In order to improve the quality of the synthesized visible images, the proposed single generator utilizes a multi-scale output architecture. Specifically, the generator $G$ produces multiple outputs at different resolution scales as follows
	\begin{equation}\label{multi-output}
	G(\mathbf{x},\mathbf{z}) = \{\mathbf{\hat{y}}_{1},\cdots,\mathbf{\hat{y}}_{s} \}, 
	\end{equation}
	where $\mathbf{x}, \mathbf{z}$ denote the input thermal image and the extracted visual-attributes, respectively.  Here,  $\{\mathbf{\hat{y}}_{1},\cdots,\mathbf{\hat{y}}_{s}\}$ denote the synthesized images with gradually growing resolutions and $\mathbf{\hat{y}}_{s}$ is the final output with the highest resolution $s$.
	In this work, we set $s=3$ where $\hat{y}_3$ is the $256\times 256$ image, $\hat{y}_2$ is the $128\times 128$ image, and $\hat{y}_1$ is the $64\times 64$ image. These multi-scale resolution outputs  act as a regularizer to the generator $G$.  Furthermore, they shorten the error signal flow path and help to improve the training stability \cite{zhang2018photographic}.

	The multi-scale generator network, as shown in Fig.~\ref{fig:generator}(a), consists of the following components: \\
	CL(64)-CNL(128)-CNL(256)-CNL(512)-CNL(512)-CNL(512)-CNL(512)-MCB(512)-DNR(512)-DNR(512)-DNR(512)-DNR(256)-DNR(128)-DNR(64)-DNR(32),\\
	where C stands for the convolutional layer (conv), L stands for LeakyReLU layer (negative\_slope=0.02), N stands for the spectral normalization layer \cite{miyato2018spectral},  MCB indicates the Multimodal Compact Bilinear module \cite{fukui2016multimodal,gao2016compact}, D stands for the deconvolutional layer (dconv), and R corresponds to the ReLU layer. All the numbers in parenthesis indicate the  channel number of the output feature maps.  Table~\ref{tab:generator structure} gives the details of the generator architecture.  Note that, for simplicity, spectral normalization \cite{miyato2018spectral}, LeakyReLU and ReLU layers are omitted.  In the last three layers, feature maps are converted into three-channel images by a ``to\_rgb" block, which  consists of one convolutional layer (parameters are indicated in  parenthesis) followed by a Tanh layer.

	\begin{table*}
		\centering
		\caption{Architecture details corresponding to the generator network.}
		\label{tab:generator structure}
		\begin{adjustbox}{max width=\textwidth}
			\begin{tabular}{|l|ccccccccccccccc|}
				\hline
				& conv & conv & conv & conv & conv  & conv & conv & MCB & dconv & dconv & dconv & dconv & \makecell{dconv \\ (to\_rgb)} & \makecell{dconv \\ (to\_rgb)} & \makecell{dconv \\ (to\_rgb)} \\
				\hline
				Input Size & 256 & 128 & 64 & 32 & 16 & 8 & 4 & 2 & 2 & 4 & 8 & 16 & 32 (128) & 64 (64) & 128 (32)\\
				Output Channel & 64 & 128 & 256 & 512 & 512 & 512 & 512 & 512 & 512 & 512 & 512 & 256 & 128 (3) & 64 (3) & 32 (3) \\
				Kernel Size & 3 & 3 & 3 & 3 & 3 & 3 & 3 & - & 3 & 3 & 3 & 3 & 3 (3) & 3 (3) & 3 (3) \\
				Stride Size & 2 & 2 & 2 & 2 & 2 & 2 & 2 & - & 2 & 2 & 2 & 2 & 2 (1) & 2 (1) & 2 (1) \\
				\hline
			\end{tabular} 
		\end{adjustbox}
	\end{table*}
	
	\subsection{Discriminator} \label{sec: discriminator}
	A series of distinct discriminators $D_{i}, i=1,\cdots, s$ are utilized and trained iteratively with the generator $G$.  For a certain discriminator at the $i$-th resolution scale, a patch-based discriminator \cite{pix2pix2017} is leveraged and it not only aims to discriminate between real/fake images but also to discriminate between the image and the corresponding attributes. Similar to previous works  \cite{reed2016generative,zhang2018photographic,StackGAN++}, a triplet of paired image and attribute is given to the discriminator: $real$, $fake$ and $wrong$. The $real$ pair consists of a real-image ($\mathbf{y}_{i}$) along with the corresponding true-attributes ($\mathbf{z}$). The $wrong$ pair consists of a real image ($\mathbf{y}_{i}$) along with wrong attributes ($\mathbf{z'}$). The $fake$ pair consists of a fake-image ($\mathbf{\hat{y}}_{i}$) with true attributes ($\mathbf{z}$). The overall adversarial objective function used to train the network is as follows:
	\begin{equation} \label{eq: multi-scale adversarial loss}
	\begin{split}
	\mathcal{L}_{G} = \sum_{i=1}^{s} \min_{G} \max_{D_{i}} (V_{real}^{i} + V_{fake}^{i} + V_{wrong}^{i}),
	\hspace{6mm} \\
	V_{real}^{i} = \mathbb{E}_{\mathbf{y}_{i}\sim P_{Y}}[\log D_{i}(\mathbf{y}_{i})] \hspace{28mm} \\ + \mathbb{E}_{\mathbf{y}_{i},\mathbf{z} \sim P_{Y,Z}}[\log D_{i}(\mathbf{y}_{i}, \mathbf{z})] \hspace{18mm}
	\\
	V_{wrong}^{i} = \mathbb{E}_{\mathbf{y}_{i}, \mathbf{z'}\sim P_{Y,Z}}[\log (1-D_{i}(\mathbf{y}_{i}, \mathbf{z'}))] \hspace{8mm}
	\\
	V_{fake}^{i} = \mathbb{E}_{\mathbf{\hat{y}}_{i} \sim P_{G(\mathbf{x},\mathbf{z})}}[\log (1-D_{i}(\mathbf{\hat{y}}_{i}))]  \hspace{14mm}
	\\ + \mathbb{E}_{\mathbf{\hat{y}}_{i} \sim P_{G(\mathbf{x},\mathbf{z})}, \mathbf{z}\sim P_{Z}}[\log (1-D_{i}(\mathbf{\hat{y}}_{i},\mathbf{z}))].
	\end{split}
	\end{equation}
	Specifically, each discriminator $D_{i}$ has two streams: conditional stream and unconditional stream. One discriminator on $256\times256$ resolution scale is illustrated in Fig.~\ref{fig:discriminator}. The unconditional stream aims to learn the discrimination between the real and the synthesized images. This unconditional adversarial loss is back-propagated to $G$ to make sure that the generated samples are as realistic as possible. In addition, the conditional stream aims to learn whether the given image matches the given attributes or not. This conditional adversarial loss is back-propagated to $G$ so that it generates samples that are attribute-preserving.

	Fig.~\ref{fig:discriminator} gives an overview of a discriminator at $256\times256$ resolution scale. This discriminator consists of 6 convolutional blocks for both conditional and unconditional streams. Details of these convolutional blocks are as follows: \\
	CL(64)-CNL(128)-CNL(256)-CNL(512)-C\footnote[2]{unconditional and conditional streams are shortened for brevity.}NL(512)-C\footnotemark[2]S(1),\\
	where S stands for the Sigmoid activation layer. Note that the only difference between the unconditional and conditional stream is the concatenation of the attribute vector at the fifth convolutional block.    
	For different discriminator, $D_{i}$ at different resolution scale, the number of convolutional down-sample blocks (blocks with dotted lines in Fig.~\ref{fig:discriminator}) vary, but we keep the bottleneck feature map at the same size (i.e. $8 \times 8$).
	The architecture details corresponding to the other discriminators are given in Table~\ref{tab:discriminator structure}.

	\begin{table}
		\centering
		\caption{Architecture details corresponding to different discriminators. Numbers in parenthesis indicate the  channel number of the output feature maps. The convolutional layers have stride size 2.}
		\label{tab:discriminator structure}
		\begin{tabular}{|p{25mm}|p{25mm}|p{25mm}|}
			\hline 
			Discriminator 64x64 & Discriminator 128x128 & Discriminator 256x256 \\ 
			\hline 
			Convolutional (64) \newline LeakyReLU  & Convolutional (64) \newline LeakyReLU & Convolutional (64) \newline LeakyReLU \\ 
			\hline 
			Convolutional (128) \newline Spectral Norm \newline LeakyReLU & Convolutional (128) \newline Spectral Norm \newline LeakyReLU & Convolutional (128) \newline Spectral Norm \newline LeakyReLU \\ 
			\hline 
			Convolutional\footnotemark[2] (256) \newline Spectral Norm \newline LeakyReLU & Convolutional (256) \newline Spectral Norm \newline LeakyReLU & Convolutional (256) \newline Spectral Norm \newline LeakyReLU \\ 
			\hline 
			Convolutional\footnotemark[2] (1) \newline Sigmoid & Convolutional\footnotemark[2] (512) \newline Spectral Norm \newline LeakyReLU  & Convolutional (512) \newline Spectral Norm \newline LeakyReLU \\ 
			\hline 
			& Convolutional\footnotemark[2] (1)  \newline Sigmoid & Convolutional\footnotemark[2] (512) \newline Spectral Norm \newline LeakyReLU \\ 
			\hline 
			&  & Convolutional\footnotemark[2] (1)  \newline Sigmoid \\ 
			\hline 
		\end{tabular} 
	\end{table}
	
	\subsection{Loss Function} \label{sec: stretagy}
	
	The generator is optimized by minimizing the following loss
	\begin{equation} \label{eq: all loss}
	\mathcal{L}_{Multi-AP-GAN} = \mathcal{L}_{G} + \lambda_{A} \mathcal{L}_{A} + \lambda_{P}\mathcal{L}_{P} + \lambda_{I}\mathcal{L}_{I} + \lambda_{1}\mathcal{L}_{1},
	\end{equation}
	where $\mathcal{L}_{G}$ is the multi-scale adversarial loss in Eq~\eqref{eq: multi-scale adversarial loss} , $\mathcal{L}_{P}$ is the perceptual loss, $\mathcal{L}_{I}$ is the identity loss, $\mathcal{L}_{A}$ is the attribute loss,
	$\mathcal{L}_{1}$ is the loss based on
	the $L_{1}$-norm between the target and the reconstructed image,  and $\lambda_{P}, \lambda_{I}, \lambda_{A}, \lambda_{1}$ are the corresponding weights. 
	
	\subsubsection{Multi-scale Perceptual and Identity Loss}
	Perceptual loss was originally introduced by Johnson \etal \cite{johnson2016perceptual} for style transfer and super-resolution.  It has been observed that the perceptual loss produces visually pleasing results than $L_{1}$ or $L_{2}$ loss.  The perceptual and identity losses are defined as follows
	\begin{equation}\label{perceptual, identity loss}
	\mathcal{L}_{P,I} = \sum_{i=1}^{s} \sum_{c=1}^{3}\sum_{w=1}^{W}\sum_{h=1}^{H}\|F(\mathbf{\hat{y}}_{i})^{c,w,h}-F(\mathbf{y}_{i})^{c,w,h}\|_{1},
	\end{equation}
	where $F$ represents a non-linear CNN feature. VGG-16 \cite{simonyan2014very} is used to extract features in this work. $C,W,H$ are the dimensions of features from a certain level of the VGG-16, which are different for perceptual and identity losses. Since the deeper convolutional layer captures more semantic information,  we choose deeper convolutional feature maps as the identity loss.
	
	In addition, multi-scale $L_{1}$ loss between the synthesized image $\mathbf{\hat{y}}_{i}$ and the corresponding real image $\mathbf{y}_{i}$ is used to capture the low-frequency information, which is defined as follows 
	\begin{equation} \label{eq:L1 loss}
	\mathcal{L}_{1} = \sum_{i=1}^{s} \| \mathbf{\hat{y}}_{i} - \mathbf{y}_{i} \|_{1}.
	\end{equation}

	\subsubsection{Multi-scale Attribute Loss}
	Inspired by the perceptual loss, we define an attribute preserving loss, which measures the error between the attributes of the synthesized image and the real image. To make sure the pre-trained model captures the facial attribute information, we fine-tune the pre-trained VGG-Face network on the annotated attribute dataset and regard the fine-tuned attribute classifier as the pre-trained model for the attribute preserving loss. Similar to the perceptual loss, the  $\mathcal{L}_A$ is defined as follows 
	\begin{equation}\label{eq:attribute loss}
	\mathcal{L}_{A} = \sum_{i=1}^{s} \|Q(\mathbf{\hat{y}}_{i})-Q(\mathbf{y}_{i})\|_{1},
	\end{equation}
	where $Q$ is the fine-tuned attribute predictor network. The output vectors are from the last layer. As a result the feature dimensions $C,W,H$ are omitted in \eqref{eq:attribute loss}. By feeding such attribute information into the generator during training, the generator $G$ is able to learn semantic information corresponding to the face.

	\subsection{Implementation} \label{sec: implementation}
	The entire network is trained in Pytorch on a single Nvidia Titan-X GPU. During the Multi-AP-GAN training, the $L_{1}$, perceptual and identity loss parameters are chosen as  $\lambda_{1}=10$, $\lambda_{P}=2.5$, $\lambda_{I}=0.5$, respectively. The ADAM \cite{kingma2014adam} is implemented as the optimization algorithm with parameter $betas=(0.5, 0.999)$ and batch size is set equal to 1. The total epochs are 200. For the first 100 epochs, we fix the learning rate as $0.0002$ and for the remaining 100 epochs, the learning rate was decreased by $1/100$ after each epoch. The feature maps for the perceptual and the identity loss are from the relu1-1 and the relu2-2 layers, respectively.  In order to fine-tune the attribute predictor network, we manually annotate images with the attributes tabulated in Table~\ref{tb:attributes table}.
	
	\begin{table}[htp!]
		\centering
		\caption{The facial attributes used in this work.}
		\begin{tabular}{|c|c|}
			\hline attributes & \makecell{Arched\_Eyebrows, Big\_Lips, Big\_Nose,\\
				Bushy\_Eyebrows, Male, Mustache,  \\ Narrow\_Eyes, No\_Beard, \\Mouth\_Slightly\_Open, Young}  \\ 
			\hline 
		\end{tabular} 
		\label{tb:attributes table}
	\end{table}

	\section{Datasets and Protocols} \label{sec: dataset}
	In this section, we describe the datasets and the protocols that we use to conduct experiments.  In particular, we describe the new extended ARL Polarimetric thermal face dataset and the corresponding protocol that we use in this paper.
	
	\subsection{Extended Polarimetric Thermal Face Dataset}
\begin{figure}[htp!]
		\centering
		\includegraphics[width=0.75\linewidth]{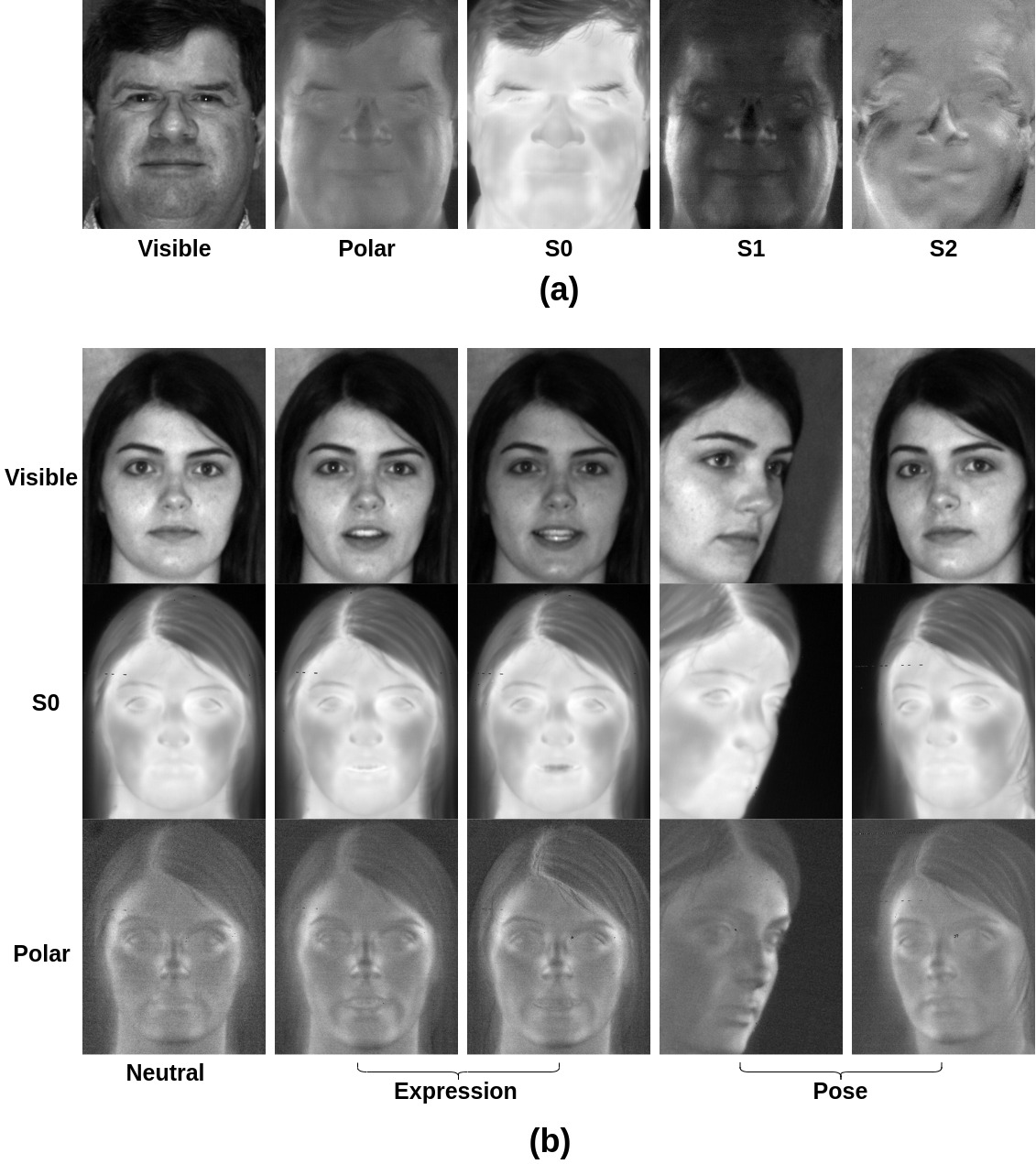}
		\caption{Sample images from the ARL dataset.  (a) Visible, polarimetric thermal, and Stokes images (S0, S1, S2) corresponding to a subject from the ARL dataset \cite{hu2016polarimetric}.  (b) Sample visible, conventional thermal and polarimetric thermal images with different variations  from the ARL Dataset Volume \RNum{3}. }
		\label{fig:datasample}
	\end{figure}

	In many recent approaches, the polarization-state information of thermal emissions has been used to achieve improved cross-spectrum face recognition performance \cite{hu2016polarimetric,riggan2016estimation,short2015improving,zhang2017generative,Riggan2018thermal} since it captures geometric and textural details of faces that are not present in the conventional thermal facial images \cite{short2015improving,hu2016polarimetric}.    A polarimetric thermal image consists of three Strokes images: $S_{0}$, $S_{1}$, $S_{2}$ where $S_{0}$ indicates the conventional total intensity thermal image, $S_{1}$ captures the horizontal and vertical polarization-state information, $S_{2}$ captures the diagonal polarization-state information \cite{hu2016polarimetric}. Similar to \cite{zhang2017generative, Riggan2018thermal}, we also refer to Polar as the three channel polarimetric image concatenated with $S_0$ , $S_1$ and $S_2$.  These Stokes images along with the visible and the polarimetric images corresponding to a subject in the ARL dataset \cite{hu2016polarimetric} are shown in Fig.~\ref{fig:datasample}. It can be observed that $S_{1}$, $S_{2}$ tend to preserve more textural details compared to $S_0$.

	The U.S. Army DEVCOM Army Research Laboratory (ARL) multimodal face dataset consists of polarimetric thermal and visible face image pairs in three volumes.  Volume \RNum{1} consists of the polarimetric thermal and visible images from 60  subjects, which were collected by the U.S. Army Research Laboratory in 2014-2015. Frontal imagery with different ranges and expressions are included. Details regarding this volume can be found in  \cite{hu2016polarimetric} and \cite{zhang2018synthesis}. Volume \RNum{2} consists of images from 51 subjects  collected at a Department of Homeland Security test facility.  As described in \cite{zhang2018synthesis}, while  the  participants  of  the Volume \RNum{1} subset consisted exclusively of the ARL employees, the participants of the Volume \RNum{2} collect were recruited from the local community in Maryland, resulting in more demographic diversity. In addition, frontal imagery with various expressions is included in this volume.
	
	In this paper, we present an extension of the dataset which was collected by ARL across 11 different sessions over 6 days. We refer to this extended dataset as Volume \RNum{3} hereinafter. Volume \RNum{3} contains polarimetric thermal and visible facial signatures from 121 subjects collected at Johns Hopkins University Applied Physics Laboratory as part of an IARPA government testing event.  There are a total of 5419 polarimetric thermal and visible image pairs with significant variations (Fig.~\ref{fig:datasample}) such as expression, off-pose, glasses, etc.  These variations make the dataset more challenging for cross-modal face verification. Note that this extended database is available upon request.
	
	
	To  be  consistent  with  previous  methods \cite{zhang2018synthesis,hu2016polarimetric},  
	the experimental protocols are defined as follows: 
	
	\noindent \textbf{Protocol \RNum{1}: }The Protocol \RNum{1} is evaluated on Volume \RNum{1}, which consists of frontal imagery with range and expression variations (including neutral expression).  Images from 30 subjects with eight samples for each subject are used as the training split. Images from the other 30 subjects with eight samples for each subject are used as the test split.  All the samples in training and test split are randomly chosen from  60 subjects.  Results are evaluated on five random splits.  Note that there are no overlapping subjects between training and test splits.  
	
	\noindent \textbf{Protocol \RNum{2}:} The Protocol \RNum{2} is evaluated on the extended 111 subject dataset which contains the images from both Volume \RNum{1} and Volume \RNum{2}.  In particular, 85-subject images are used as the training split and the other 26-subject images are denoted as the test split. The 85-subject images in training split consist of all 60-subject images in Volume \RNum{1} and another 25-subject images randomly selected from Volume \RNum{2}. The other 26-subject images in Volume \RNum{2} are selected as the test split. As before, results are evaluated on five random splits \cite{zhang2018synthesis}.  Note that Volume \RNum{2} consists of frontal imagery with expression variations only (including neutral expression).

	\noindent \textbf{Protocol \RNum{3}:} The Protocol \RNum{3} is evaluated only on the Volume \RNum{3} data consisting of images from 121 subjects.  Volume \RNum{3} includes frontal and off-pose imagery (excludes extreme pose, e.g. profile), and expression variation (including neutral expression).  Images from 96 randomly chosen subjects are used as the training split and the images from the remaining 25 subjects are used as the test split. Results are evaluated on five random splits. 
	
	\subsection{Visible and Thermal Paired Face Database}
	In addition to the ARL dataset, the proposed method is evaluated on a recently introduced Visible and Thermal Paired Face Database \cite{Mallat18}. This dataset contains thermal and visible image pairs corresponding to 50 subjects.   Each subject participated in two different sessions separated by a time interval of 3 to 4 months.  This dataset includes 21 face images per subject in each session.  These images correspond to different facial variations in illumination, head pose, expression and occlusion. In total, 4200 images are included in this dataset.

	\noindent \textbf{Protocol: }  Images corresponding to randomly chosen 30 subjects are used as the training split and the images from the remaining 20 subjects are used as the test split. This results in 630 paired training images and 420 paired testing images.  There is no overlap among subjects in the training and the test sets.  Results are evaluated on five random splits.

	\subsection{Tufts Face Database}
\begin{figure}[t!]
		\centering
		\includegraphics[width=0.8\linewidth]{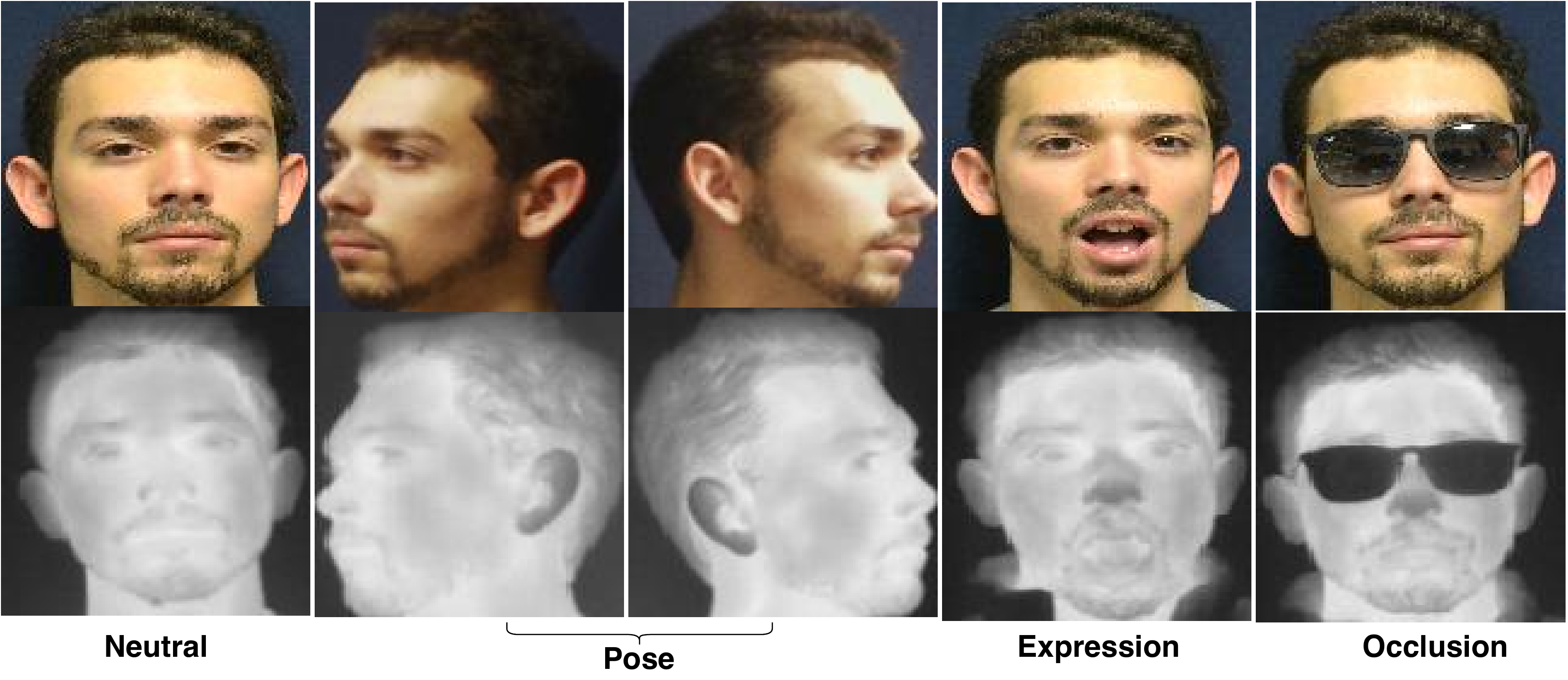}
		\caption{Sample thermal and visible images from the Tufts Face Database \cite{TUFSFaceDataset} with different variations.}
		\label{fig:datasample_tufs}
	\end{figure}

	We also evaluate the proposed method on a recently proposed Tufts Face Database \cite{TUFSFaceDataset}, which contains 1532 paired visible and thermal face images from 112 subjects.  For each subject multiple images are taken in different conditions.  In particular, each subject has images in 9 different poses, 4 expressions and 1 occlusion with eye glasses.  Sample images from this dataset are shown in \ref{fig:datasample_tufs}.  The Tufts dataset \cite{TUFSFaceDataset} is more difficult than the other two datasets as it contains less number of images per person in each variation.

	
	\noindent \textbf{Protocol: } Similar to the previous protocols, images corresponding to 90 subjects are used for training and the images from the remaining 22 subjects are used for testing. This results in about 1232 paired data for training and 300 paired data for testing.  There is no overlap among subjects in the training and the test sets.  Results are reported based the evaluations on five random splits.

	\subsection{Preprocessing}
	In addition to the standard preprocessing, two more preprocessing steps are used for the proposed method. First, the faces in the visible images are detected by MTCNN \cite{mtcnn}. Then, a standard central crop method is used to crop the detected faces. Since MTCNN is implementable on the visible images only, we use the same detected rectangle coordinations to crop the thermal images, which were already aligned to the same canonical coordinates as the visible images. After preprocessing, all the images are scaled and saved as  $256\times256$ 16-bit PNG files.

	\subsection{Metrics}
	Once the visible image is synthesized from the input probe thermal image, we use a pre-trained VGG-Face model \cite{parkhi2015deep} to extract features from the synthesized visible probe image as well as the visible gallery image to perform cross-modal face verification.  In particular, the verification score is calculated using the cosine similarity between the two feature vectors.  The cross-modal verification performance of different methods is evaluated using the Receiver Operating Characteristic (ROC) curve, Area Under the Curve (AUC) and Equal Error Rate (EER) measures.

	\begin{figure*}[t]
		\centering
		\begin{minipage}{.48\textwidth}
			\centering
			\includegraphics[width=0.9\linewidth]{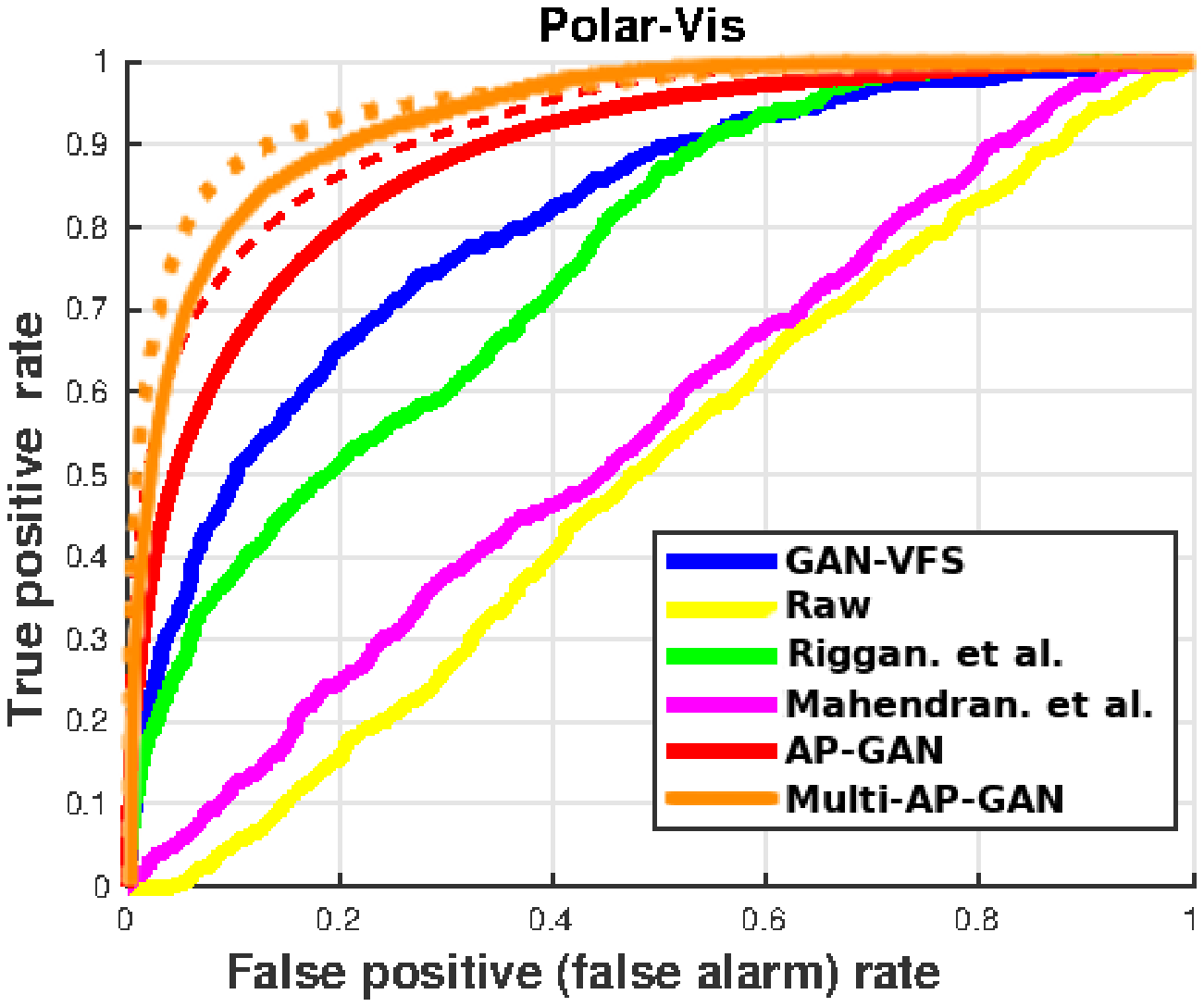}\\
			(a) Polarimetric thermal-to-visible verification 
		\end{minipage}
		\begin{minipage}{.48\textwidth}
			\centering
			\includegraphics[width=0.9\linewidth]{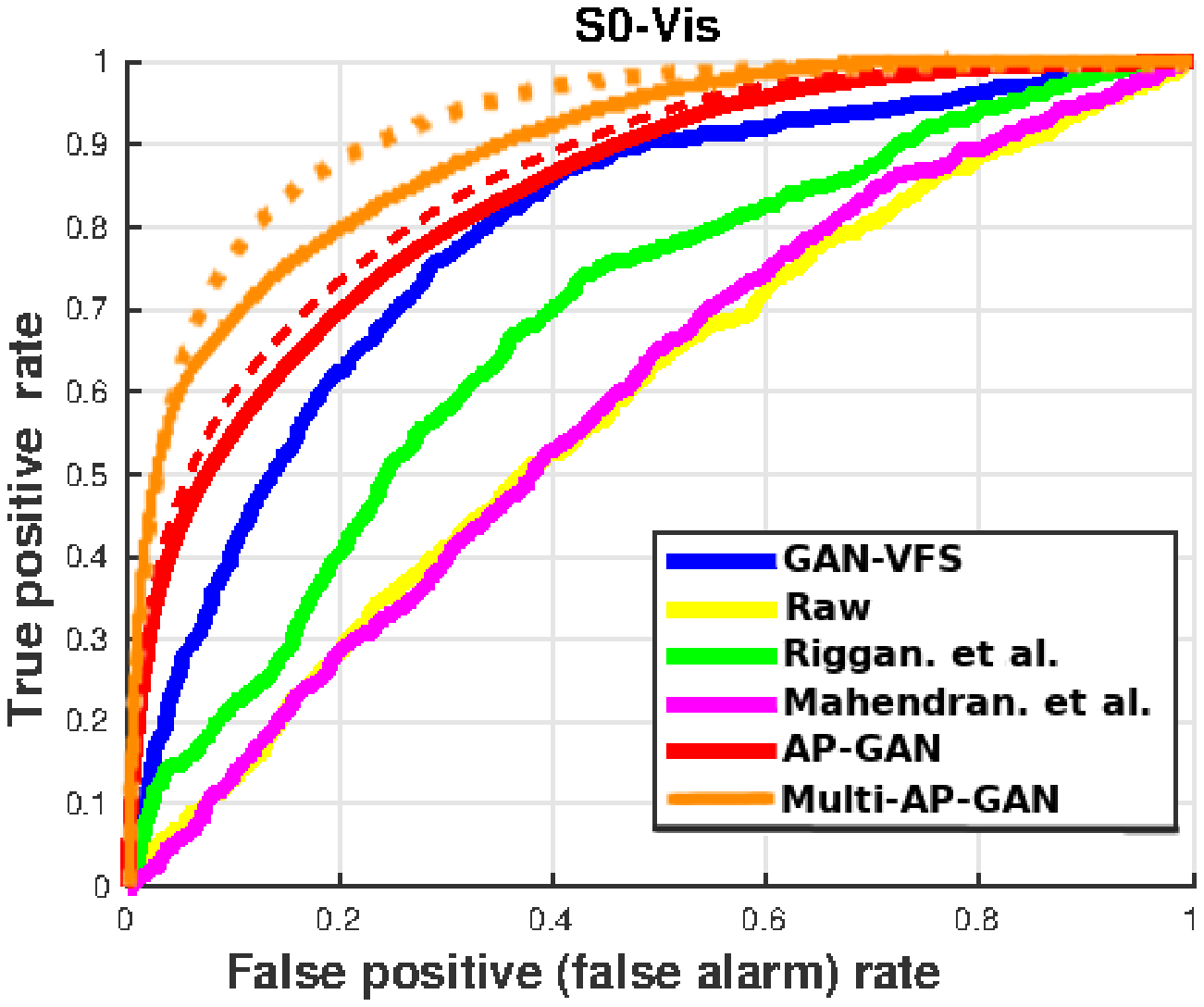}\\
			(b) Thermal-to-visible verification \hspace{5mm}
		\end{minipage}
		\caption{The ROC curve comparison on Protocol \RNum{1} with several state-of-the-art methods: (a) Polarimetric thermal-to-visible verification performance. (b) S0-to-Visible verification performance. Note that the dotted lines indicate results based on the ground-truth attributes. The gap between the results with ground-truth attributes and that with predicted attributes demonstrate the degradation caused by the attribute predictor.}
		\label{fig:Comparison_Figure}
	\end{figure*}
	
	\begin{figure*}[t]
		\centering
		\begin{minipage}{.48\textwidth}
			\centering
			\includegraphics[width=0.9\linewidth]{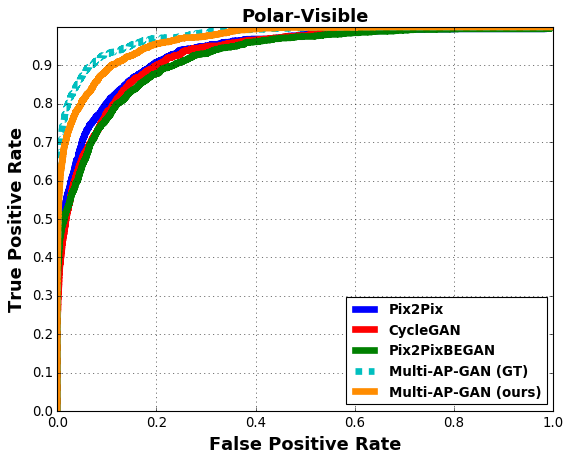}\\
			(a) Polarimetric thermal-to-visible verification 
		\end{minipage}
		\begin{minipage}{.48\textwidth}
			\centering
			\includegraphics[width=0.9\linewidth]{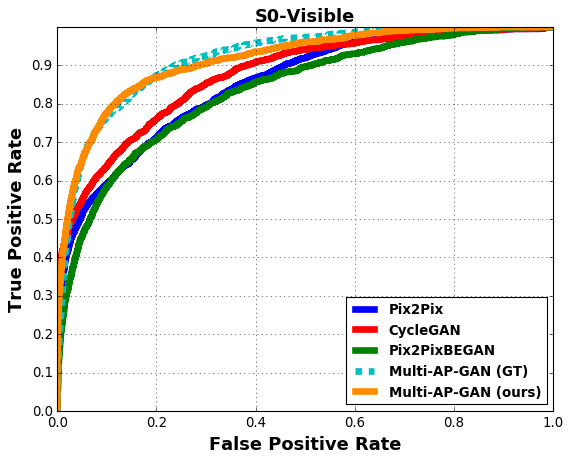}\\
			(b) Thermal-to-visible verification \hspace{5mm}
		\end{minipage}
		\caption{The ROC curve comparison on Protocol \RNum{2} with several state-of-the-art methods: (a) Polarimetric thermal-to-visible verification performance. (b) S0-to-Visible verification performance. Note that the dotted lines indicate results based on the ground-truth attributes. The gap between the results with ground-truth attributes and that with predicted attributes demonstrate the degradation caused by the attribute predictor.}
		\label{fig:Protocol2_Comparison_Figure}
	\end{figure*}

	\begin{figure*}[t]
		\centering
		\begin{minipage}{.48\textwidth}
			\centering
			\includegraphics[width=0.9\linewidth]{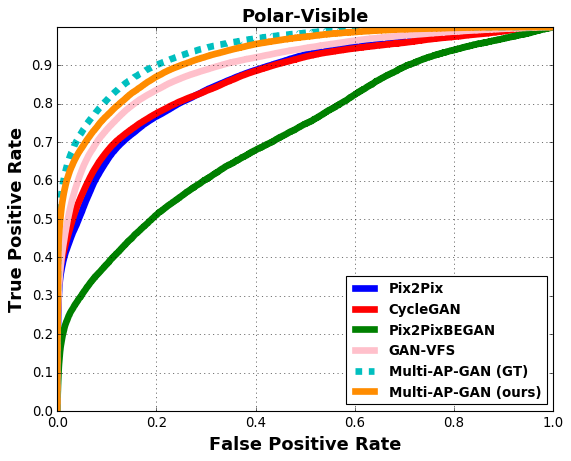}\\
			(a) Polarimetric thermal-to-visible verification 
		\end{minipage}
		\begin{minipage}{.48\textwidth}
			\centering
			\includegraphics[width=0.9\linewidth]{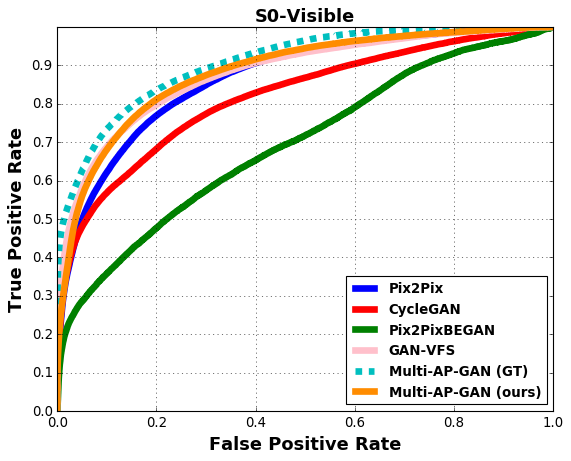}\\
			(b) Thermal-to-visible verification \hspace{5mm}
		\end{minipage}
		\caption{The ROC curve comparison on Protocol \RNum{3} with several state-of-the-art methods: (a) Polarimetric thermal-to-visible verification performance. (b) S0-to-Visible verification performance. Note that the dotted lines indicate results based on the ground-truth attributes. The gap between the results with ground-truth attributes and that with predicted attributes demonstrate the degradation caused by the attribute predictor.}
		\label{fig:Protocol3_Comparison_Figure}
	\end{figure*}
	
	\begin{figure}[t]
		\centering
		\includegraphics[width=0.85\linewidth]{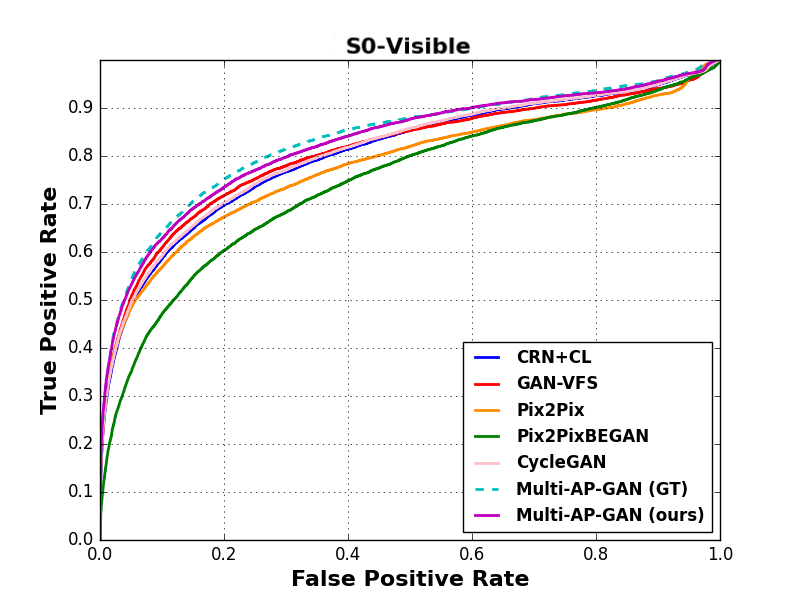}
		\caption{The ROC curve comparison on Thermal-Visible Paaired Database \cite{Mallat18}. Note that the dotted lines indicate results based on the ground-truth attributes. Similarly, the gap between the results with ground-truth attributes and that with predicted attributes demonstrate the degradation caused by the attribute predictor. }
		\label{fig:thvis_roc.png}
	\end{figure}	
	
	\begin{figure}[t]
		\centering
		\includegraphics[width=1\linewidth]{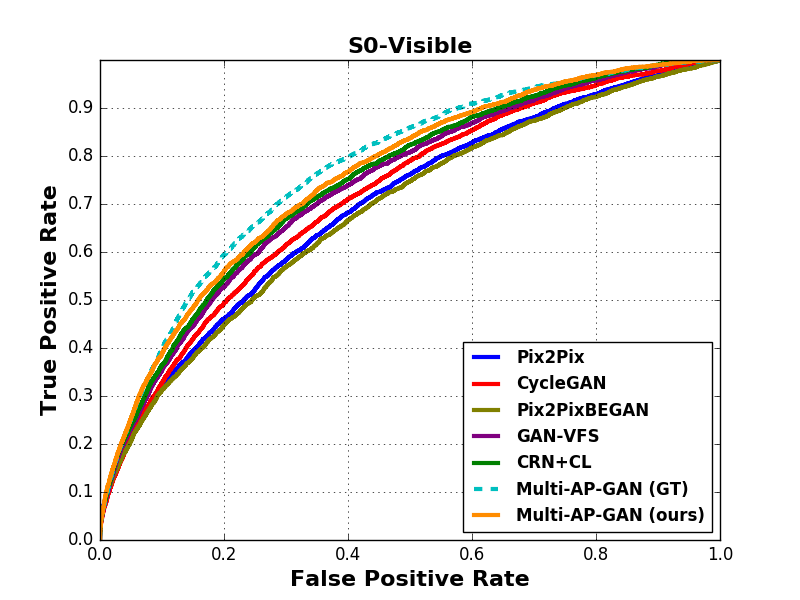}
		\caption{The ROC curve comparison on the Tufts Face Database \cite{TUFSFaceDataset}. Note that the dotted lines indicate results based on the ground-truth attributes. Similarly, the gap between the results with the ground-truth attributes and that with predicted attributes demonstrate the degradation caused by the attribute predictor. }
		\label{fig:tufs_roc.png}
	\end{figure}
	
	\begin{figure*}[t]
		\centering
		\includegraphics[width=0.95\linewidth]{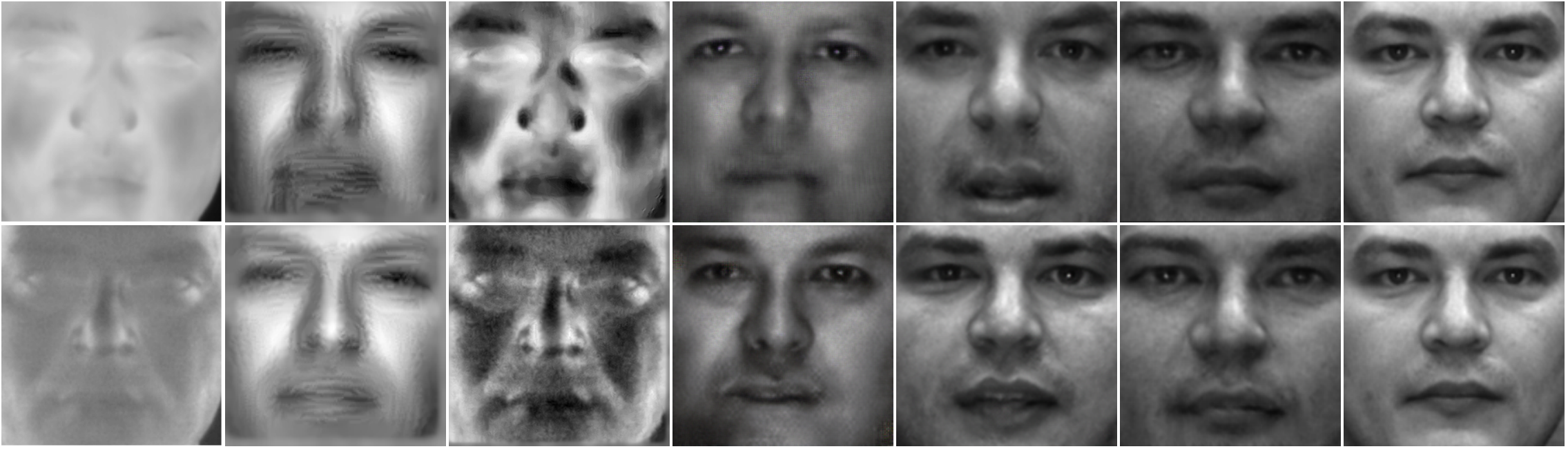}\\
		\raggedright
		\footnotesize
		\hspace{13mm} Input \hspace{10mm} Riggan \etal \cite{riggan2016estimation} \hspace{1mm} Mahendran \etal \cite{mahendran2015understanding} \hspace{2mm} GAN-VFS \cite{zhang2017generative}\hspace{6mm} AP-GAN \cite{di2018polarimetric}  \hspace{6mm} Multi-AP-GAN \hspace{6mm} Ground Truth\\
		\caption{The visual comparison of synthesized samples from different methods: Riggan \etal \cite{riggan2016estimation}, Mahendran \etal \cite{mahendran2015understanding}, GAN-VFS \cite{zhang2017generative},  AP-GAN \cite{di2018polarimetric}, Multi-AP-GAN, Ground Truth. The first row results correspond to the S0 image, and the second row results correspond to the Polar image.}
		\label{fig:comparison_figure}
	\end{figure*}

	\begin{figure*}[t]
		\centering
		\includegraphics[width=0.95\linewidth]{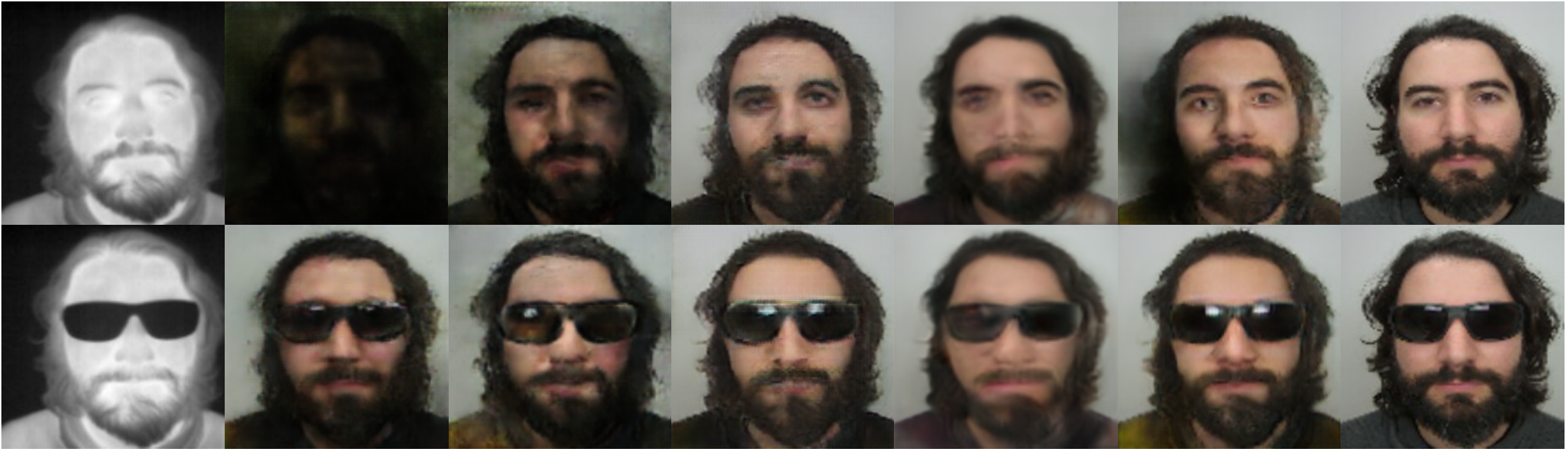}\\
		\raggedright
		\small
		\hspace{15mm} Input \hspace{10mm} Pix2Pix \cite{pix2pix2017} \hspace{4mm} CycleGAN \cite{CycleGAN2017} \hspace{2mm} GAN-VFS \cite{zhang2017generative}\hspace{2mm} CRN+CL\scalebox{0.8}{ \cite{mallat2019cross,damer2019cascaded}}  \hspace{2mm} Multi-AP-GAN \hspace{2mm} Ground Truth \\
		\caption{The visual comparison of synthesized images corresponding to Pix2Pix\cite{pix2pix2017}, CycleGAN \cite{CycleGAN2017}, GAN-VFS \cite{zhang2017generative}, CRN+CL \cite{mallat2019cross,damer2019cascaded}, Multi-AP-GAN (ours) from the Visible and Thermal Paired Face Database \cite{Mallat18}.}
		\label{fig:vis_th_samples}
	\end{figure*}

	\begin{figure*}
		\centering
		\includegraphics[width=0.95\linewidth]{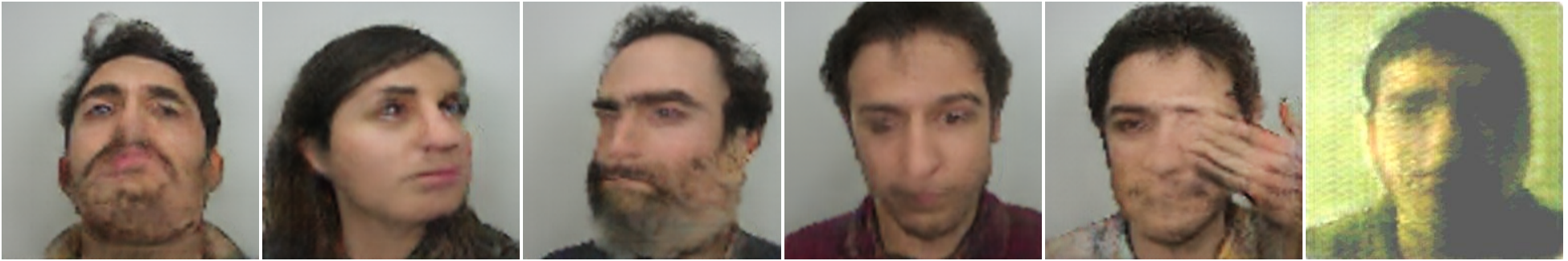}\\
		\raggedright
		\small
		\hspace{15mm} Pose-Up \hspace{12mm} Pose-Left \hspace{15mm} Pose-Right \hspace{12mm} Pose-Down \hspace{2mm} Occolusion-Hand on Eye \hspace{4mm} Light-Dark \\
		\caption{Some failure cases. Note that extreme pose, illumination and occlusion variations cause the proposed method to synthesize poor quality images.}
		\label{fig:vis_th_failure}
	\end{figure*}

	\begin{figure*}[t]
		\centering
		\includegraphics[width=0.95\linewidth]{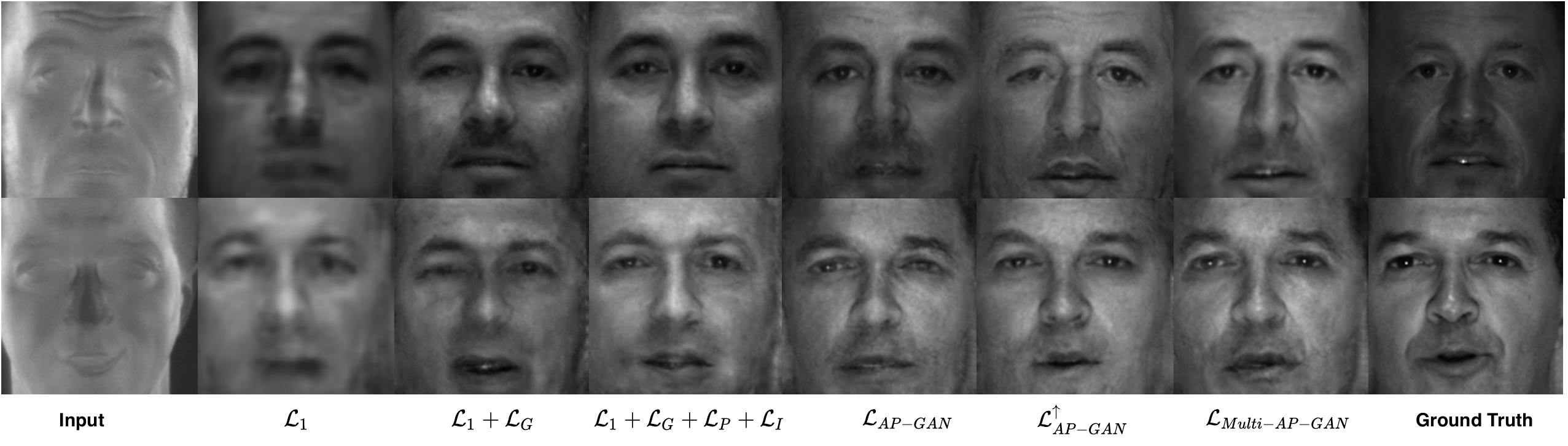}\\
		\caption{The visual results of the ablation study for different experimental settings. Given input polarimetric thermal image, synthesized results using different combination of losses and resolutions are shown successively from left to right. One intermediate synthesis results $\mathcal{L}^{\uparrow}_{AP-GAN}$, which utilizes 2-scale resolutions $128\times 128$ and $256\times 256$, is shown here to demonstrate the progressive improvements obtained by adding multi-scale.}
		\label{fig:ablation_figure}
	\end{figure*}
	
	\section{Experimental Results} \label{sec: experimental result}
	In this section, we demonstrate the effectiveness of the proposed approach by conducting various experiments on the datasets described in the previous section.  Since the ARL Dataset contains both conventional thermal $(S_{0})$ and polarimetric thermal modalities, we conduct the following two cross-modal face verification experiments on the ARL dataset: 1) Conventional thermal (S0) to Visible (Vis) and 2) Polarimetric thermal (Polar) to Visible (Vis).  On the other hand, the Visible and Thermal Paired Face Database  and the Tufts Face Database do not contain polarimetric thermal images.  As a result, we only conduct thermal-to-visible cross-domain face verification experiments on these datasets.  
	
	We evaluate and compare the performance of the proposed method with that of the following recent state-of-the-art methods \cite{zhang2017generative,mahendran2015understanding,riggan2016estimation,Riggan2018thermal,di2018polarimetric,zhang2018synthesis,mallat2019cross,damer2019cascaded}.   Note that our previous work \cite{di2018polarimetric} can be viewed as a single scale version of the proposed method. In particular, in \cite{di2018polarimetric}, we synthesize images at a particular scale which has the same resolution as the input.  We also conduct experiments with another baseline method called, Multi-AP-GAN (GT), where we use the ground-truth attributes in our method rather than automatically predicting them using the proposed attribute predictor. This baseline will clearly determine how effective the proposed attribute predictor is in determining the attributes from unconstrained visible faces.

	\subsection{Results on the ARL Face Dataset}	
	Fig.~\ref{fig:Comparison_Figure} shows the performance corresponding to 
	Protocol \RNum{1} on two different experimental settings (i.e S0-to-visible and Polar-to-visible). Compared with other state-of-the-art methods in Fig.~\ref{fig:Comparison_Figure}, the proposed method performs better with a larger AUC and lower EER scores. In addition, it can be observed that the performance corresponding to the Polar modality is better than the S0 modality, which also demonstrates the advantage of using the polarimetric thermal images than the conventional thermal images.  In addition, the gap between the results with ground-truth attributes (dash-line) and that with the predicted attributes (solid-line) demonstrates the degradation caused by the attribute predictor. The quantitative comparisons, as shown in the Table~\ref{tb: comparison_table}, also demonstrate the effectiveness of the proposed method. In addition, compared with the previous single scale resolution method \cite{di2018polarimetric}, the proposed multi-scale algorithm achieves significant improvement: around $4\%$  and $6\%$ on the conventional and polarimetric thermal modalities, respectively. These improvements demonstrate the effectiveness of the proposed multi-scale synthesis algorithm. 
	
	Furthermore, we also show some visual comparisons in Fig.~\ref{fig:comparison_figure}. The first row in Fig.~\ref{fig:comparison_figure} shows one synthesized sample using S0. The second row shows the same synthesized sample using a polarimetric thermal image. It can be observed that the results of  Riggan \etal \cite{riggan2016estimation} do capture the overall face structure but it tends to lose some facial details. Results of Mahendran \etal \cite{mahendran2015understanding} are poor compared to \cite{riggan2016estimation}.  Results of Zhang \etal \cite{zhang2017generative} are more photo-realistic but tend to lose some attribute information. The proposed Multi-AP-GAN not only generates photo-realistic images but also preserves attributes on the reconstructed images.  
	
	Fig.~\ref{fig:Protocol2_Comparison_Figure} and Table~\ref{tb:protocol 2} show the performance of different methods on Protocol \RNum{2}. These results also demonstrate the superiority of the proposed method.  Note that the performance of many methods is slightly better in Protocol \RNum{2} than Protocol \RNum{1}.  This is mainly due to the fact that  the training dataset is larger in Protocol \RNum{2} than Protocol \RNum{1}.

	Protocol \RNum{3} results corresponding to different methods are shown in Fig.~\ref{fig:Protocol3_Comparison_Figure} and Table~\ref{tb:arl121}. Note that face images in this volume include many variations such as expression, pose, illuminations and occlusion (glasses).  As a result, the performance of the methods compared is slightly lower than what we observed in  Protocol \RNum{1}  and Protocol \RNum{2}.  In general, the proposed method performs favorably against the state-of-the-art methods.  Note that Pix2PixBEGAN method \cite{pix2pix2017,berthelot2017began} fails to generate good quality visible faces from profile thermal face images.  As a result, Pix2PixBEGAN method performs poorly on this dataset.

	We further analyze the cross-modal verification performance of different methods on different variation settings on Protocol \RNum{3}.  The corresponding results are shown in  Table~\ref{tb:arl121_variation}.  Since variations like occlusion and illumination are not included in some subjects, we only use three variations (neutral, expression, and pose) which are included in all subjects.  As can be seen from Table~\ref{tb:arl121_variation}, the performance degradation mainly comes from pose variations.

	\subsection{Results on the Visible and Thermal Paired Face Database}
	Table~\ref{tb:eurocom} shows the  performance of different methods on the Visible and Thermal Paired Face Database.  Compared to the ARL Face dataset, the performance of every method is lower on this dataset.  This is mainly due to the fact that this dataset is small in size and contains many facial variations.  In general, the proposed method performs favorably against the previous methods.

	In addition, following the analysis presented in \cite{Mallat18}, we also analyze how different variations (i.e. illumination, pose, expression, occlusion) influence the cross-spectrum matching performance of our method.  As can be seen from the results in Table~\ref{tb:eurocom_variation} illumination and pose variations are the two variations that affect the performance of our method the most.   This analysis is based on the proposed method implemented with the ground-truth visual attributes.
	
	We also show some visual results in Fig.~\ref{fig:vis_th_samples}.  It can be observed that Pix2Pix \cite{pix2pix2017} and CycleGAN \cite{CycleGAN2017}  methods generate poor quality images with many artifacts.  GAN-VFS \etal \cite{zhang2017generative} is able to synthesize better quality images.   However, this method also introduces some artifacts around the eyes and mouth regions.   The proposed Multi-AP-GAN method not only generates photo-realistic images but also preserves attributes on the synthesized images.  We also show some images in Fig.~\ref{fig:vis_th_failure} in which the proposed method is not able to produce good quality images. From these images we see that extreme pose, occlusion and illumination variations cause the proposed method to produce poor quality images.   
	
	\subsection{Results on the Tufts Face Database}
	Table~\ref{tb:tufs} and Fig. \ref{fig:tufs_roc.png} show the  performance of different methods on the Tufts Face Dataset \cite{TUFSFaceDataset}.  Compared to the previous two datasets, this dataset is more challenging due to a large number of pose and expression variations as well as a few number of images per variation, which leads to the lower performance of every method. In general, our method outperforms the other baseline methods on this challenging dataset by improvements on 1.8 \% EER and 2.4\% AUC scores respectively.
	

	\begin{table*}[t]
		\caption{ARL Protocol \RNum{1} verification performance comparisons among the baseline methods, state-of-the-art methods, and the proposed Multi-AP-GAN method for both polarimetric thermal (Polar) and conventional thermal (S0) cases.}
		\centering
		\begin{tabular}{|c|c|c|c|c|}
			\hline Method & AUC(Polar) & AUC(S0) & EER(Polar) & EER(S0) \\ 
			\hline Raw & $50.35\%$ & $58.64\%$ & $48.96\%$ & $43.96\%$ \\ 
			\hline Mahendran \etal \cite{mahendran2015understanding} & $58.38\%$ & $59.25\%$ & $44.56\%$ & $43.56\%$ \\ 
			\hline Riggan \etal \cite{riggan2016estimation}  & $75.83\%$  & $68.52\%$ & $33.20\%$ & $34.36\%$ \\ 
			\hline GAN-VFS \etal \cite{zhang2017generative} & $79.90\%$  & $79.30\%$  & $25.17\%$  & $27.34\%$ \\ 
			\hline Riggan \etal \cite{Riggan2018thermal} & $85.43\%$ & $82.49\%$ & $21.46\%$ & $26.25\%$ \\ 
			\hline AP-GAN \cite{di2018polarimetric} & $88.93\%\pm 1.54\%$ & $84.16\%\pm 1.54\%$ & $19.02\%\pm 1.69\%$ & $23.90\%\pm 1.52\%$\\ 
			\hline AP-GAN (GT) \cite{di2018polarimetric} & $91.28\%\pm 1.68\%$ & $86.08\%\pm 2.68\%$ & $17.58\%\pm 2.36\%$ & $23.13\%\pm 3.02\%$ \\
			\hline Multi-stream GAN  \cite{zhang2018synthesis} &$\mathbf{96.03}$\% &85.74\%&$11.78$\%&23.18\%\\
			\hline Multi-AP-GAN (ours) & $93.61\%\pm 1.46\%$ & $\mathbf{90.14\% \pm 2.17\%}$ & $14.24\%\pm 1.91\%$ & $\mathbf{18.20\% \pm 2.65\%}$ \\        
			\hline Multi-AP-GAN (GT) (ours) & $95.29\%\pm 1.39\%$ & $\mathbf{92.72\% \pm 2.03\%}$ & $\mathbf{11.22\%\pm 1.89}\%$ & $\mathbf{16.05\% \pm 2.15\%}$ \\ 
			\hline 
		\end{tabular}
		\label{tb: comparison_table}
	\end{table*}

	\begin{table*}[htp!]
		\centering
		\caption{ARL Protocol \RNum{2} verification performance comparisons among the baseline methods and the proposed method for both polarimetric thermal (Polar) and conventional thermal (S0) cases.}
		\begin{tabular}{|c|c|c|c|c|}
			\hline 
			Method & AUC (Polar) & AUC(S0) & EER(Polar) & EER(S0) \\ 
			\hline 
			Raw & 66.85\% & 63.66\% & 37.85\% & 40.93\% \\ 
			\hline
			Pix2Pix \cite{pix2pix2017} & $93.66\% \pm 1.07\%$ & $85.09\% \pm 1.48\%$ & $13.73\% \pm 1.38\%$ & $23.12\% \pm 1.14\%$  \\
			\hline
			Pix2PixBEGAN \cite{pix2pix2017,berthelot2017began} & $92.16\% \pm 1.09\%$ & $83.69\% \pm 1.28\%$& $15.38\% \pm 1.45\%$ & $26.22\% \pm 1.16\%$  \\
			\hline
			CycleGAN \cite{CycleGAN2017} (supervised)  &  $93.11\% \pm 1.02\% $ & $87.29\% \pm 1.13\% $  & $15.19\% \pm 1.02\%$  & $20.99\% \pm 1.19\%$  \\ 
			\hline 
			Multi-stream GAN  \cite{zhang2018synthesis} &$\mathbf{98.00}$\% &--&$7.99$\%&--\\
			\hline 
			Multi-AP-GAN (ours) & $96.55\%\pm 1.12\%$ & $\mathbf{91.43\% \pm 0.93\%}$ & $10.17\%\pm 1.01\%$ & $\mathbf{15.86\% \pm 2.13\%}$ \\
			
			\hline 
			Multi-AP-GAN (GT) (ours) & $97.68\%\pm 0.78\%$ & $\mathbf{91.88\% \pm 0.87\%}$ & $\mathbf{7.69\%\pm 1.39\%}$ & $\mathbf{15.29\% \pm 2.36\%}$ \\
			\hline    
		\end{tabular}
		\label{tb:protocol 2}
		\vspace{-2mm}
	\end{table*}
	
	\begin{table*}[htp!]
		\begin{minipage}{0.95\textwidth}\centering
			\caption{ARL Protocol \RNum{3}  verification performance comparisons among the baseline methods and the proposed method for both polarimetric thermal (Polar) and conventional thermal (S0) cases.}    
			
			\begin{tabular}{|c|c|c|c|c|}
				\hline 
				Method & AUC (Polar) & AUC(S0) & EER(Polar) & EER(S0) \\ 
				\hline 
				Raw & 73.43\% & 76.71\% & 33.56\% & 30.76\% \\ 
				\hline
				Pix2Pix \cite{pix2pix2017} &  $86.78\% \pm 1.84\%$ & $86.65\% \pm 1.48\%$ & $21.92\% \pm 1.26\%$ & $23.12\% \pm 1.77\%$  \\
				\hline
				Pix2PixBEGAN \cite{pix2pix2017,berthelot2017began} & $71.29\% \pm 1.88\%$ & $69.42\% \pm 1.84\%$& $33.83\% \pm 1.68\%$ & $36.88\% \pm 1.76\%$  \\
				\hline
				CycleGAN \cite{CycleGAN2017} (supervised)  &  $86.77\% \pm 1.77\% $ & $81.80\% \pm 1.67\% $  & $21.48\% \pm 1.11\%$  & $25.86\% \pm 1.36\%$  \\ 
				\hline 
				GAN-VFS \footnotemark[3] \cite{zhang2017generative} & $90.20\% \pm 1.85\%$  & $87.10\% \pm 1.52\%$ & $18.53\% \pm 1.21\%$ & $20.22\% \pm 1.92\%$ \\
				\hline 
				Multi-AP-GAN (ours) & $\mathbf{92.29\%\pm 1.48\%}$ & $\mathbf{88.49\% \pm 1.87\%}$  & $\mathbf{16.26\% \pm 1.12\%}$ & $\mathbf{19.25\% \pm 1.62\%}$  \\
				\hline 
				Multi-AP-GAN (GT) (ours) & $\mathbf{93.72\%\pm 1.08\%}$ & $\mathbf{90.99\% \pm 1.13\%}$ & $\mathbf{14.75\%\pm 1.36\%}$ & $\mathbf{17.81\% \pm 1.63\%}$ \\
				\hline    
			\end{tabular}
			\label{tb:arl121}
		\end{minipage}
		
		\begin{minipage}{0.95\textwidth}\centering
			\caption{Protocol \RNum{3} verification performance with respect to different variations.}
			\begin{tabular}{|c|c|c|c|c|}
				
				\hline 
				Variations & AUC (Polar) & AUC(S0) & EER(Polar) & EER(S0) \\ 
				\hline 
				Neutral & $96.77\% \pm 1.25\%$ & $94.69\% \pm 1.17\%$ & $12.50\% \pm 2.09\% $ & $13.38\% \pm 1.48\% $\\
				\hline 
				Expression & $96.77\% \pm 1.91\%$ & $92.38\% \pm 1.40\%$ & $10.05\% \pm 2.02\% $ & $15.18\% \pm 1.58\% $ \\ 
				\hline
				Pose  & $86.62\% \pm 2.39\%$ & $82.35\% \pm 2.54\%$ & $22.45\% \pm 1.84\% $ & $25.76\% \pm 1.95\% $ \\
				\hline
				Average  & $93.72\%\pm 1.08\%$ & $90.99\% \pm 1.13\%$ & $14.75\%\pm 1.36\%$ & $17.81\% \pm 1.63\%$ \\
				\hline 
			\end{tabular}
			\label{tb:arl121_variation}
			\vspace{-2mm}
		\end{minipage}
	\end{table*}

	\begin{table*}[htp!]
		\centering
		
		\begin{minipage}{0.48\textwidth}\centering
			\caption{Visible and Thermal Paired Face Database verification performance comparisons among the baseline methods and the proposed method for the conventional thermal case.}    
			\begin{tabular}{|c|c|c|}
				\hline 
				Method & AUC & EER \\ 
				\hline 
				Raw & 69.54\% & 35.39\%  \\ 
				\hline
				Pix2Pix \cite{pix2pix2017} & $78.66\% \pm 1.48\%$ & $28.39\% \pm 1.14\%$  \\
				\hline
				Pix2PixBEGAN \cite{pix2pix2017,berthelot2017began} & $73.69\% \pm 1.82\%$ & $34.22\% \pm 1.61\%$  \\
				\hline
				CycleGAN \cite{CycleGAN2017} (supervised)  & $80.24\% \pm 1.31\% $  & $26.72\% \pm 1.39\%$  \\
				\hline
				GAN-VFS \footnotemark[3] \cite{zhang2017generative} & $80.44\% \pm 1.03\% $  & $26.33\% \pm 1.19\%$  \\
				\hline  
				CRN + CL \footnotemark[3] \cite{mallat2019cross,damer2019cascaded} & $81.25\% \pm 1.01\% $  & $26.01\% \pm 1.23\% $  \\
				\hline 
				Multi-AP-GAN (ours)  & $\mathbf{81.73\% \pm 0.93\%}$ & $\mathbf{25.68\% \pm 1.56\%}$ \\
				\hline 
				Multi-AP-GAN (GT) (ours)  & $\mathbf{82.68\% \pm 0.87\%}$ & $\mathbf{23.16\% \pm 0.98\%}$ \\
				\hline    
			\end{tabular}
			\label{tb:eurocom}
		\end{minipage}
		\hfill
		\begin{minipage}{0.48\textwidth}\centering
			\caption{Verification performance with respect to different variations on the Visible and Thermal Paired Face Database.}
			\begin{tabular}{|c|c|c|}
				
				\hline 
				Variations & AUC & EER \\ 
				\hline 
				Illumination & $73.35\% \pm 0.25\%$ & $32.60\% \pm 0.43\%$ \\
				\hline 
				Expression & $97.25\% \pm 0.68\%$ & $7.45\% \pm 1.74\%$  \\ 
				\hline
				Pose & $78.25\% \pm 1.03\%$ & $28.75\% \pm 0.93\%$  \\
				\hline
				Occlusion & $83.98\% \pm 1.33\%$ & $24.02\% \pm 1.06\%$  \\
				\hline
				Average  & $82.68\% \pm 0.87\%$  & $23.16\% \pm 0.98\%$  \\
				\hline 
			\end{tabular}
			\label{tb:eurocom_variation}
			\vspace{-2mm}
		\end{minipage}
	\end{table*}

	\begin{figure*}
		\centering
		\includegraphics[width=0.95\linewidth]{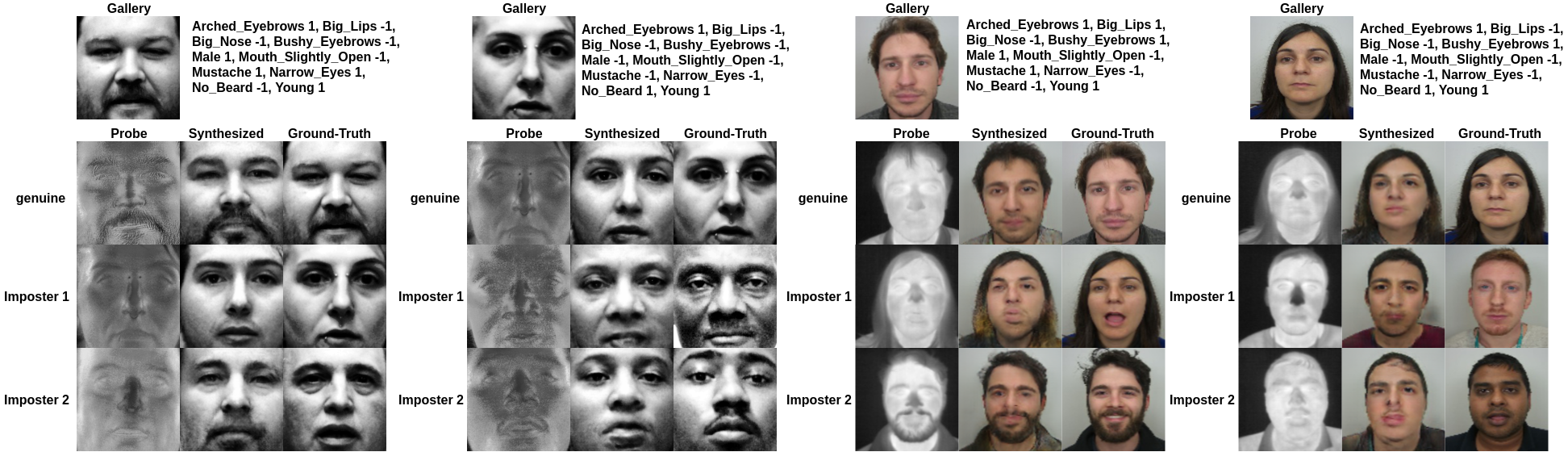}
		\caption{Analysis of attributes on synthesis. We show the synthesis samples from either conventional or polarimetric thermal images on both datasets. Given probe (thermal) images and estimated attributes from the gallery (visible) image, our proposed method can generates attribute preserving (visible) images.}
		\label{fig:verificationsample}
	\end{figure*}

	\begin{table}[htp!]
		\centering
		\caption[]{The Tufts Face Database \cite{TUFSFaceDataset} verification performance comparisons among the baseline methods and the proposed method.}
		\label{tb:tufs}
		\resizebox{.48\textwidth}{!}{    
			\begin{tabular}{|c|c|c|}
				\hline 
				Method & AUC & EER \\ 
				\hline 
				Raw & 66.73\% & 38.13\%  \\ 
				\hline
				Pix2Pix \cite{pix2pix2017} & $69.73\% \pm 0.92\%$ & $35.83\% \pm 0.59\%$  \\
				\hline
				Pix2PixBEGAN \cite{pix2pix2017,berthelot2017began} & $68.89\% \pm 0.51\%$ & $36.88\% \pm 0.43\%$  \\
				\hline
				CycleGAN \cite{CycleGAN2017} (supervised)  & $71.93\% \pm 1.94\%$  & $34.16\% \pm 1.70\%$  \\
				\hline
				GAN-VFS \footnotemark[3] \cite{zhang2017generative} & $73.78\% \pm 0.46\% $  & $32.32\% \pm 0.53\%$  \\
				\hline  
				CRN + CL \footnotemark[3] \cite{mallat2019cross,damer2019cascaded} & $74.90\% \pm 0.56\% $  & $31.71\% \pm 0.54\%$  \\
				\hline 
				Multi-AP-GAN (ours)  &  $\mathbf{75.86\% \pm 0.88\%}$ & $\mathbf{31.14\% \pm 0.74\%}$  \\
				\hline 
				Multi-AP-GAN (GT) (ours)  & $\mathbf{77.38\% \pm 0.98\%}$ & $\mathbf{29.94\% \pm 0.79\%}$ \\
				\hline
			\end{tabular}
		}
	\end{table}

	\subsection{Ablation Study}

	\begin{figure}[t]
		\centering
		\includegraphics[width=0.85\linewidth]{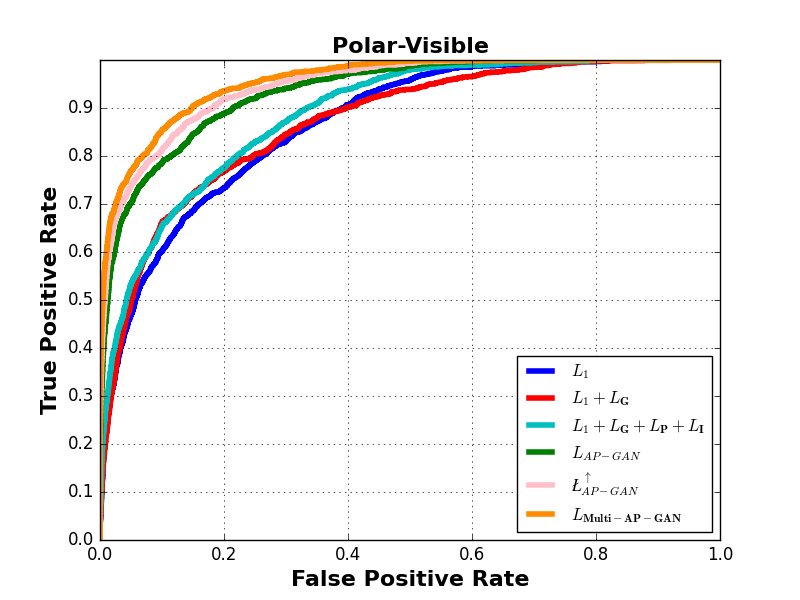}
		\caption{The ROC curves corresponding to the ablation study.}
		\label{fig:Ablation_Polar}
	\end{figure}

	In  order  to  demonstrate  the  effectiveness  of  different modules in the proposed method, we conduct the following ablation study using the Polarimetric thermal modality in the  ARL dataset on Protocol \RNum{1}:\\ 
	\begin{enumerate}[noitemsep]
		\item Polar to Visible estimation with only $\mathcal{L}_{1}$ (as defined in Eq.~\eqref{eq:L1 loss})
		\item Polar to Visible estimation with $\mathcal{L}_{1}$ and $\mathcal{L}_{G}$ (as defined in Eq.~\eqref{eq: multi-scale adversarial loss})
		\item Polar to Visible estimation with $\mathcal{L}_{1}$, $\mathcal{L}_{G}$, perceptual loss $\mathcal{L}_{P}$ and identity loss  $\mathcal{L}_{I}$, which are defined as in Eq.~\eqref{perceptual, identity loss}.
		\item Polar to Visible estimation with all the losses as defined in  Eq.~\eqref{eq: all loss}, by utilizing various solution scales: $\mathcal{L}_{AP-GAN}$ ($256^{2}$), $\mathcal{L}^{\uparrow}_{AP-GAN}$ ($128^{2}, 256^{2}$), $\mathcal{L}_{Multi-AP-GAN}$ ($64^{2}, 128^{2}, 256^{2}$) respectively.
	\end{enumerate}
	Fig.~\ref{fig:Ablation_Polar} shows the ROC curves corresponding to each experimental setting. From this figure, we can observe that using all the losses together as $\mathcal{L}_{Multi-AP-GAN}$ can obtain the best performance. Compared to the results between $\mathcal{L}_{1}$ and $\mathcal{L}_{1}+\mathcal{L}_{G}$, we can observe the enhancement provided by adding the adversarial loss. Compared with the results between $\mathcal{L}_{1}+\mathcal{L}_{G}$ and $\mathcal{L}_{1}+\mathcal{L}_{G} + \mathcal{L}_{P} + \mathcal{L}_{I}$, we can observe the improvements obtained by adding the perceptual and identity loses.  On the other hand, one can clearly see the significance of fusing the semantic attribute information with the image feature in the latent space by comparing the results between $\mathcal{L}_{1}+\mathcal{L}_{G} + \mathcal{L}_{P} + \mathcal{L}_{I}$ and $\mathcal{L}_{AP-GAN}$. Additionally, looking at the comparison with $\mathcal{L}_{AP-GAN}$, $\mathcal{L}^{\uparrow}_{AP-GAN}$ and $\mathcal{L}_{Multi-AP-GAN}$, one can see the successive improvements by leveraging the multi-scale information.

	Besides the ROC curves, we also show the visual results for each experimental setting in Fig.~\ref{fig:ablation_figure}. Given the input Polar image, the synthesized results from different experimental settings are shown in Fig.~\ref{fig:ablation_figure}. It can be observed that  $\mathcal{L}_{1}$ captures the low-frequency features of images very well. $\mathcal{L}_{1} + \mathcal{L}_{G}$ can capture both low-frequency and high-frequency features in the image. However, it adversely introduced distortions and artifacts in the synthesized image. In addition, optimizing $\mathcal{L}_{P} + \mathcal{L}_{I}$ suppresses these distortions to some extent.  Finally, fusing attributes into the loss on with leveraging multi-scale resolution (i.e. $\mathcal{L}_{Multi-AP-GAN}$) can not only improving the performance but also preserves facial attributes. 
	In our study, we do not see significant more improvement by utilizing more than 3-scale resolutions.

	
	In addition, we analyze the effect of attributes on the synthesized images in Figure~\ref{fig:verificationsample}.  In particular, given the input gallery image, we examine how attributes help in synthesizing a visible image from a thermal probe image.   If the probe image and the input gallery image share the same identity then Multi-AP-GAN is able to generate attribute preserving visible image. On the other hand, if the probe image's identity is different from that of the gallery image then  the proposed method is not able to synthesize identity preserving visible face.  However, the attributes are still preserved on the synthesized image.  This analysis further demonstrates that the proposed Multi-AP-GAN method learns the cross-spectral (thermal-to-visible) translation mapping exactly guided by the visual attributes.

	\footnotetext[3]{results are obtained after re-implementation due to the limited code availability.}
	\footnotetext[1]{features are extracted: https://github.com/TreB1eN/InsightFace\_Pytorch}
	
	\section{Discussion}
	The proposed Multi-AP-GAN approach generates better quality visible images and as a result obtains improved cross-modal verification performance compard to previous GAN-based approaches.  This can be contributed to  the fact that Multi-AP-GAN uses a better generator which is guided by visual attributes.  The multi-scale generator mitigates the  receptive-field limitation of the convolutional operation by leveraging  the features corresponding to images at multiple scales.  In addition, visual attributes provide complementary semantic information for better synthesis.  GAN-based methods such as GAN-VFS \cite{zhang2017generative}, Multi-stream GAN \cite{zhang2018synthesis} and Pix2Pix \cite{pix2pix2017} are single-scale generators and do not exploit such facial semantic information during synthesis.
	
	Though our method performs reasonably well on three datasets, there are some limitations which we hope to overcome in our future work.  Our model requires  paired thermal and visible face images for training, which is laborious and expensive.  Hence, an unsupervised synthesis method that does not require paired data  is needed.  Another limitation of our approach is that it does not work well on extreme pose variations.  We are currently developing a new method that can deal with this pose issue in heterogeneous face recognition. We also plan to further investigate the impact of metabolic and physiologic variability in thermal facial signatures on synthesis and subsequent recognition performance.


	\section{Conclusion} \label{sec: conclusion}
	We propose a novel Attribute Preserving Generative Adversarial Network (Multi-AP-GAN) structure for thermal-to-visible face verification via synthesizing photo-realistic visible face images from the corresponding thermal (polarimetric or conventional) images with extracted attributes. Rather than use only image-level information for synthesis and verification, we take a different approach in which semantic facial attribute information is also fused during training and testing. Quantitative and visual experiments evaluated on a real thermal-visible dataset demonstrate that the proposed method achieves state-of-the-art performance compared with other existing methods. In addition, an ablation study is developed to demonstrate the improvements obtained by different combination of loss functions.

	\section*{Acknowledgment}
	This  work  was  supported  by  the  Defense  Forensics  \& Biometrics  Agency  (DFBA).  The  authors  would  like  to thank Mr.  Tom Cantwell and Ms.  Michelle Giorgilli for their guidance and extensive discussions on this work.  The authors would like to express their appreciation to ODNI/IARPA, as well as Chris Nardone, Marcia Patchan, and Stergios Papadakis at the JHU Applied Physics Laboratory for enabling ARL's participation in the 2018 IARPA Odin Program data collection, and Mathew Thielke for his help in collecting the data.

	{\small
		\bibliographystyle{ieee}
		\bibliography{tbiom}
	}
	
\end{document}